\documentclass[11pt]{article}

\usepackage[final]{acl}

\usepackage{times}
\usepackage{latexsym}
\usepackage{booktabs}
\usepackage{tabularx}
\usepackage[T1]{fontenc}

\usepackage[utf8]{inputenc}
\usepackage[utf8]{inputenc}
\usepackage[T1]{fontenc}
\usepackage{hyperref}
\usepackage{url}
\usepackage{booktabs}
\usepackage{amsfonts}
\usepackage{nicefrac}
\usepackage{microtype}
\usepackage{xcolor}

\usepackage{microtype}
\usepackage{graphicx}
\usepackage{subcaption}
\usepackage[table]{xcolor}
\usepackage{booktabs}

\usepackage[most]{tcolorbox}

\usepackage{amsmath}
\usepackage{amssymb}
\usepackage{mathtools}
\usepackage{amsthm}
\usepackage{algorithm}
\usepackage{makecell}
\usepackage{algpseudocode}
\usepackage{multirow}
\usepackage{enumitem}

\usepackage[capitalize,noabbrev]{cleveref}

\theoremstyle{plain}

\theoremstyle{definition}

\theoremstyle{remark}

\newtcolorbox{promptbox}{
  colback=gray!5,
  colframe=gray!40,
  boxrule=0.5pt,
  arc=3pt,
  boxsep=4pt,
  left=6pt,
  right=6pt,
  top=4pt,
  bottom=4pt,
  fontupper=\ttfamily,
  breakable,
  enhanced
}

\usepackage[textsize=tiny]{todonotes}
\usepackage{microtype}

\usepackage{inconsolata}

\usepackage{graphicx}

\title{DICS: Exploring Data Intrinsic Consistency for Visual Instruction Selection}

\author{
  \textnormal{Yuyang Hong}$^{1,2,3,*}$ \quad
  \textnormal{Jinhui Guo}$^{3*}$ \quad
  \textnormal{Jiaqi Gu}$^{3}$ \quad
  \textnormal{Lubin Fan}$^{3,\dagger}$ \\
  Ruixiang Wang$^{1,2}$ \quad
  Kun Ding$^{1,2}$ \quad
  Yue Wu$^{3}$ \quad
  Shiming Xiang$^{1,2,\dagger}$ \quad
  Jieping Ye$^{3}$ \\
  $^{1}$School of Artificial Intelligence, University of Chinese Academy of Sciences \\
  $^{2}$MAIS, Institute of Automation, Chinese Academy of Sciences \\
  $^{3}$Alibaba Token Hub, Alibaba Group \\
  \texttt{\{hongyuyang2023@ia.ac.cn, lory.gjh@alibaba-inc.com, lubin.flb@alibaba-inc.com\}} \\
  $^{*}$These authors contributed equally to this work. \\
  $^{\dagger}$Corresponding authors.
}
\begin{document}
\maketitle
\renewcommand{\thefootnote}{}
\footnotetext{\noindent\hspace*{-1.8em}This work was done during Yuyang Hong's internship at Alibaba Group.}
\renewcommand{\thefootnote}{\arabic{footnote}}
\begin{abstract}
\label{sec:abs}

Visual instruction tuning is crucial for advancing the vision-language alignment and instruction-following capabilities of Vision-Language Models (VLMs). However, identifying optimal subsets under a fixed ratio constraint from rapidly expanding datasets remains a significant bottleneck. While existing methods largely depend on distribution diversity or heuristic filtering, they often overlook the internal coherence within individual samples. 
To bridge this gap, we propose \emph{Data Intrinsic Consistency (DIC)}, a self-scoring metric designed to quantify the sample-level inter-component consistency. DIC consists of two modules: \emph{Visual Information Consistency (VIC)}, evaluating the alignment between visual content and instructions, and \emph{Response Information Consistency (RIC)}, assessing response coherence relative to the instruction. Building upon DIC, we introduce \emph{Data Intrinsic Consistency Selection (DICS)}, an adaptive data selection method that optimizes the trade-off between high intra-sample consistency and global distributional diversity under varying data budgets. 
Extensive experiments demonstrate that DICS consistently outperforms state-of-the-art methods across diverse dataset scales and model architectures,
surpassing full-dataset fine-tuning while using only 25\% of
the LLaVA-1.5-665K data.
We further curate DICS-6M, a 6M-sample
multi-modal instruction corpus that enables the largest-scale visual
instruction selection study to date; remarkably, DICS reaches 94.52\% of the official InternVL3-8B-Instruct performance using less than 25\% of its reported training data.
Code can be seen at https://github.com/cqu-student/DICS.
\end{abstract}
\section{Introduction}
\label{sec:intro}

\begin{figure}[t]
    \centering
    \includegraphics[width=1.0\linewidth]{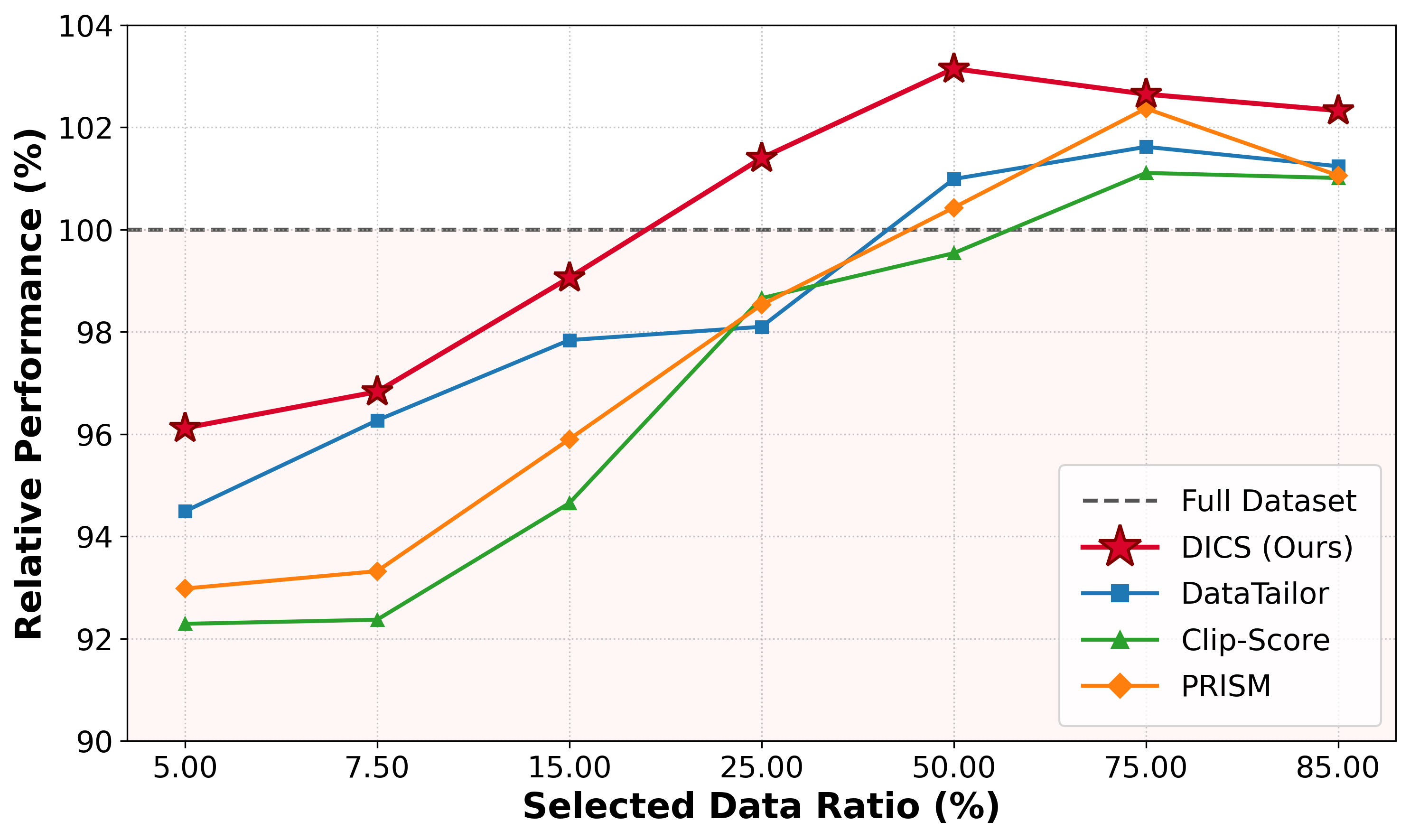}
    \caption{\textbf{Performance of LLaVA-1.5-7B fine-tuned on subsets of LLaVA-1.5-665K selected by various methods.} DICS-selected subsets outperform the full dataset at 25\% and consistently surpass other methods.}
    \label{fig:teaser}
    \vspace{-5mm}
\end{figure}

Visual instruction tuning~\cite{ouyang2022training,cui2023efficient,chiang2023vicuna} is pivotal for VLM training, establishing foundational capabilities while enabling effective human instruction understanding and following. Given the massive scale of such data, rigorous evaluation and selection are critical.

\begin{figure*}[t!]
    \centering
    \includegraphics[width=0.95\linewidth]{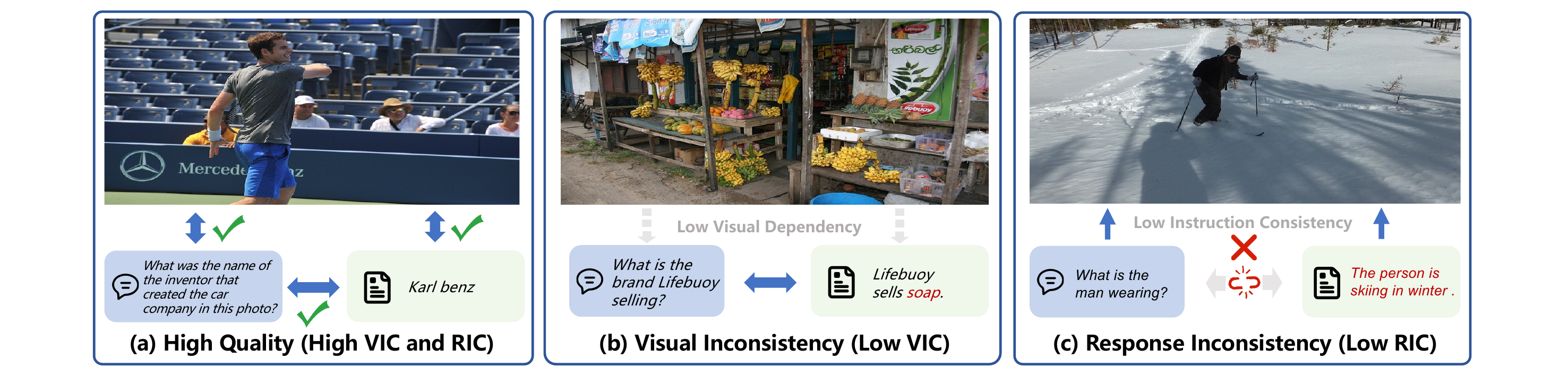}
    \vspace{-5pt}
    \caption{\textbf{Examples of Data Intrinsic Consistency in Visual Instruction Data.} (a) High-quality samples exhibiting both high VIC and RIC. (b) Low-VIC samples showing low visual dependency between cross-modal inputs and responses. (c) Low-RIC samples showing weak alignment of the response with the image and instruction. }
    \label{fig:example_incosistency}
    \vspace{-5mm}
\end{figure*}

Existing selection methods are mainly based on distribution diversity~\cite{bi2025prism}, sample relationships~\cite{yu2025mastering}, importance~\cite{wu2024icons}, and heuristic quality rules~\cite{saada2025data,li2023reflection} to identify compact subsets, removing redundancy to achieve superior VLM performance. However, these methods neglect the intrinsic relationships among images, instructions, and responses within individual samples. While similar efforts exist for language data~\cite{li2024quantity}, they are inapplicable to multimodal settings due to the lack of mechanisms to evaluate visual information's impact on quality. 

In this work, we propose \textbf{Data Intrinsic Consistency (DIC)}, a sample-level self-scoring metric grounded in the hypothesis that high-quality visual instruction data must exhibit internal consistency among its image, instruction, and response components. DIC consists of two modules: \textbf{Visual Information Consistency (VIC)}, which assesses whether the image is directly relevant to the instruction-response pair and provides critical context for generating the response; and \textbf{Response Information Consistency (RIC)}, which evaluates whether the response is focused on the instruction, accurate, and logically coherent given the visual input. Fig.~\ref{fig:example_incosistency} illustrates examples of inconsistency between images, instructions, and responses in various visual instruction samples. Intuitively, low-VIC samples exhibit weak alignment between instructions and images, making them prone to hallucinations. Similarly, low-RIC samples provide responses clearly irrelevant to both the instruction and visual content, potentially disrupting the model's vision-language alignment. To unify VIC and RIC, we introduce the DIC metric via a self-scoring framework that contrasts training losses under partial versus complete instruction scenarios. A larger loss discrepancy signifies a tighter cross-modal alignment and richer learning potential, effectively filtering high-quality samples to enhance accuracy and mitigate hallucinations.

Correspondingly, we introduce \textbf{Data Intrinsic Consistency Selection (DICS)}, a data selection method that jointly considers DIC scores, data diversity, and sampling efficiency under varying budget constraints. Its core principle involves an adaptive sampling strategy: when the selection ratio is low, it prioritizes high-DIC samples while maintaining data diversity. Conversely, when the selection ratio is high, it removes low-DIC samples exhibiting redundant features. This approach ensures the consistent selection of high-quality subsets across different data scales and resource limitations.

Extensive experiments demonstrate the significant efficacy of our DICS in visual instruction tuning. As shown in Fig.~\ref{fig:teaser}, on the widely adopted LLaVA-1.5-665K dataset, our method consistently outperforms state-of-the-art baselines across all sampling ratios. Notably, a mere 25\% subset selected by our approach achieves performance comparable to full-data fine-tuning. Furthermore, as the sampling ratio increases, our method consistently surpasses full-data baselines.
To validate effectiveness and practicality on large-scale datasets, we curated \textbf{DICS-6M}, a 6M-sample corpus assembled entirely from open-source datasets following the InternVL3~\cite{zhu2025internvl3} collection protocol, and conducted the largest data selection experiment to date 

on the open-source InternVL3-8B~\cite{zhu2025internvl3}.
Our results show that using only 25\% of the data yields superior performance compared to fine-tuning on the full 6M dataset. 

Compared against the more advanced official InternVL3-8B-Instruct, our model matches 94.52\% of its performance with less than 25\% of its training data (5.1M vs.\ 21.7M).
We will open-source our trained models and datasets.
In summary, our contributions are as follows:
\begin{itemize}
    \item We propose Data Intrinsic Consistency, comprising Visual Information Consistency and Response Information Consistency, which leverages a unified evaluation paradigm to quantify sample-level intrinsic consistency among images, instructions, and responses.
    \item We design a unified DICS sampling strategy that integrates DIC scores, diversity, and sampling efficiency to achieve budget-aware optimal subset selection.
    \item Comprehensive experiments demonstrate that DICS-selected datasets achieve SOTA performance on various benchmarks, exhibiting superior robustness and scalability across varying data scales and model architectures.
\end{itemize}

\section{Related Work}
\label{sec:related}

\subsection{Visual Instruction Tuning}
\label{sec:related:tuning}

Visual instruction tuning~\cite{ouyang2022training,cui2023efficient,chiang2023vicuna} is essential for aligning Large Language Models (LLMs) with visual modalities. Following the instruction-tuning in NLP~\cite{lou2024large,tinn2023fine}, seminal works like LLaVA~\cite{liu2023visual} and MiniGPT-4~\cite{zhu2024minigpt} utilized GPT-4 generated data to bridge visual encoders with LLMs. The field has since evolved through two main trajectories: 
\textbf{(1) Data Scaling}, exemplified by ShareGPT4V~\cite{chen2024sharegpt4v} and InstructBLIP~\cite{dai2023instructblip}, which emphasize expanding dataset scale and task variety; and 
\textbf{(2) Architectural Refinement}, where models like BLIP-2~\cite{li2023blip} introduce novel bottleneck structures, and recent models like LLaVA-NeXT~\cite{li2024llava} and Qwen2-VL~\cite{Wang2024Qwen2VLEV} adopt dynamic resolution mechanisms to capture fine-grained visual details across arbitrary scales. 

\subsection{Visual Instruction Selection}
\label{sec:related:selection}

To mitigate the computational costs of training on massive datasets, recent works focus on identifying high-value subsets and generally fall into two categories: representation-based and gradient-based methods. 
\textbf{(1) Representation-based methods}: These approaches involve filtering data based on global distributional properties to reduce redundancy. Methods like DataTailor~\cite{yu2025mastering} and PRISM~\cite{bi2025prism} utilize embedding-based clustering to ensure concept coverage and select representative samples. Other approaches~\cite{tang2025middo,hong2025taming} further optimize for distinctiveness within the feature space. Additionally, ARDS~\cite{yang2025data} constructs robust training mixtures by prioritizing samples semantically close to worst-case evaluation subgroups through clustering and perturbations. These methods are effective at maintaining diversity and ensuring the selected subset covers the original distribution.
\textbf{(2) Gradient-based methods}: These methods assess data value through training dynamics. LESS~\cite{xia2024less} computes low-rank gradient similarity to select data that benefits specific target tasks. Similarly, COINCIDE~\cite{lee2024concept}, ICONS~\cite{wu2024icons} and TIVE~\cite{liu2024less} employ influence functions to identify samples that consistently contribute to model generalization or task difficulty. These methods focus on the relational influence of each data sample to maximize transferability.
Although the aforementioned methods effectively reduce dataset size, they share a common limitation: they primarily focus on global diversity or distributional coverage, while largely neglecting the intrinsic quality of individual samples. Furthermore, while recent works~\cite{chen2024we, lee2026selective} emphasize image-text consistency, they either rely on expensive external LLMs for sample-level filtering, or only consider unidirectional alignment between modalities, both of which are unscalable for million-level training data selection.

In this paper, we propose Data Intrinsic Consistency (DIC) to assess sample-level alignment and integrate it into DICS, a unified framework that prunes inconsistent samples while preserving distributional diversity for robust training. Unlike text-only methods like IFD~\cite{li2024quantity} that cannot evaluate visual contributions, our DICS explicitly quantifies visual information gain by contrasting image-conditioned and image-free scenarios. Furthermore, while relevance-based metrics like CLIP-Score~\cite{hessel2021clipscore} conflate semantic correlation with necessity, DICS isolates the indispensable visual information, filtering out samples reliant on language priors.

\section{Data Intrinsic Consistency}
\label{sec:dic}

\begin{figure*}
    \centering
    \includegraphics[width=1\linewidth]{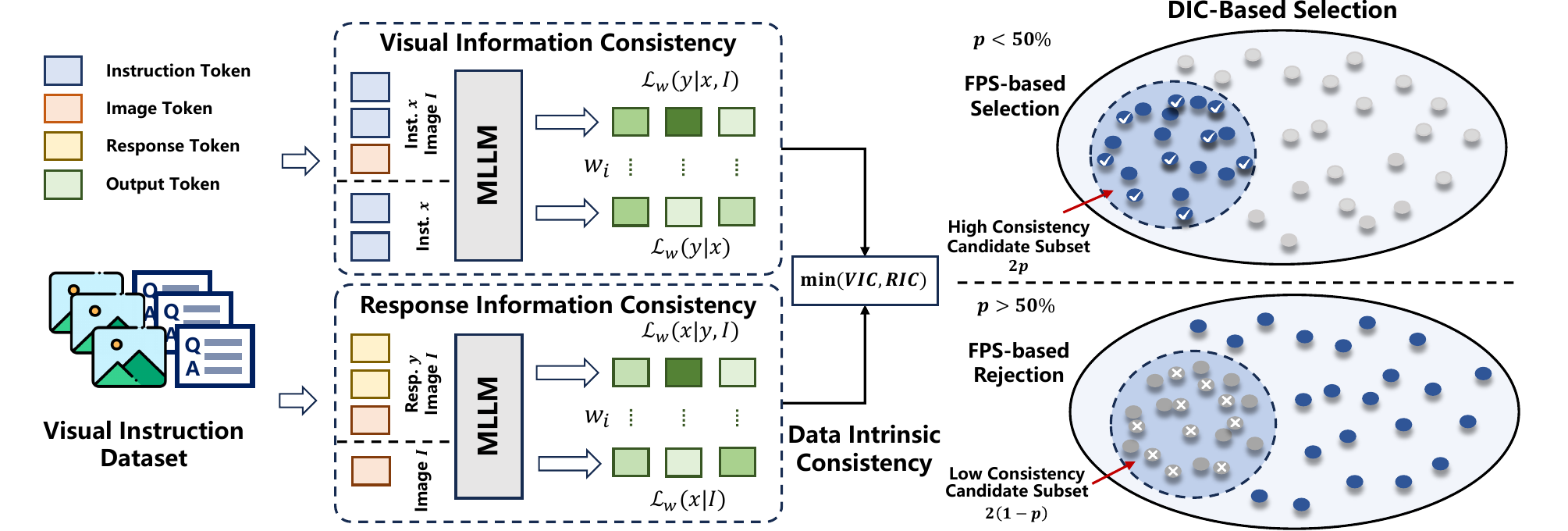}
    \caption{\textbf{Overview of Data Intrinsic Consistency Selection (DICS).} First, we compute the Visual Information Consistency and Response Information Consistency for each sample to derive its Data Intrinsic Consistency (DIC) score. Subsequently, leveraging these DIC scores and a target sampling ratio, we employ a designed adaptive sampling strategy to construct a dataset characterized by high intrinsic consistency and preserved diversity.}
    \label{fig:pipeline}
    \vspace{-5mm}
\end{figure*}

A visual instruction data sample $(I, x, y)$ consists of three components: image $I$, instruction $x$, and response $y$. 
It plays a pivotal role in fine-tuning VLMs by establishing alignment between vision and language, thereby enhancing visual instruction following capabilities. 
A high-quality instruction sample should exhibit semantic coherence and mutual information gain among $I$, $x$, and $y$. Specifically, instruction $x$ and response $y$ must be grounded in the visual content of the image $I$, ensuring that the generated response requires a substantive understanding of the visual input rather than relying solely on language priors. Furthermore, the response $y$ should provide accurate and rich content in information while maintaining strong alignment with both the instruction $x$ and the image $I$.
To this end, we propose \textbf{Data Intrinsic Consistency (DIC)}, which evaluates sample quality from two perspectives: Visual Information Consistency (VIC) and Response Information Consistency (RIC).

\subsection{Visual Information Consistency}
\label{sec:dic:vic}

Since instructions and responses that are semantically inconsistent with visual content can disrupt the vision-language alignment established during pre-training, we draw inspiration from the principle of information gain~\cite{Shannon1948AMT} and prior work~\cite{li2024quantity}. We quantify VIC by measuring the reduction in predictive uncertainty after observing the image. Specifically, we compute the difference in the model's loss when predicting the response $y$ with and without the image $I$:
\begin{equation}
    \text{VIC}(I,x,y) = \exp\!\left( \mathcal{L}_w(y \mid x) - \mathcal{L}_w(y \mid x, I) \right),
\end{equation}
where $\mathcal{L}_w(\cdot)$ denotes the cross-entropy loss weighted by token importance. The $\exp(\cdot)$ maps the loss difference into the positive domain. A higher VIC score indicates that the image provides substantial information gain for response generation. 
To boost semantic sensitivity, we employ a POS-based token weighting strategy to suppress neutral tokens (e.g., prepositions). The weighted loss is defined as:
\begin{equation}
    \mathcal{L}_w(y \mid c) = -\frac{\sum_{i=1}^{N} w_i \log p(t_i \mid c, t_{<i})}{\sum_{i=1}^{N} w_i},
\end{equation}
where $N$, $t_i$, $t_{<i}$, and $c$ denote the token count, current token, preceding context, and conditioning input ($x$ or $(x, I)$), respectively; and $w_i$ represents the POS-derived weight. Implementation details are in Appendix~\ref{sec:appendix:token_weight}.
Then the VIC score can be expressed as:
\begin{equation}
\small
\begin{split}
    \text{VIC}(I,x,y) = \exp\Bigg(&\frac{1}{\sum_{i=1}^{N} w_i} \sum_{i=1}^{N} w_i \Big[ -\log p(t_i \mid x, t_{<i}) \\
    &+ \log p(t_i \mid x, t_{<i}, I) \Big]\Bigg).
\end{split}
\label{eq:vic}
\end{equation}
Samples with responses strongly grounded in visual content receive higher VIC scores, whereas vision-irrelevant or hallucinatory samples score lower. For multi-turn dialogues, the VIC score is averaged across all turns.

\subsection{Response Information Consistency}
\label{sec:dic:ric}

To evaluate the consistency between response $y$ and instruction $x$, we introduce Response Information Consistency (RIC). It assesses if a response is specifically tailored to the query by framing it as an ``inverse prediction'' task: conditioned on visual context $I$, we measure how well $y$ aids reconstructing the original instruction $x$. 
Specifically, we employ carefully designed prompts (Appendix~\ref{sec:appendix:ric_impl}) to guide the model in inferring the original instruction from the response. For clarity, we omit the explicit prompt tokens and define the RIC as:
\begin{equation}
   \text{RIC}(I,x,y) = \exp\!\left( \mathcal{L}_w(x \mid I) - \mathcal{L}_w(x \mid y, I) \right),
\end{equation}
where $\mathcal{L}_w(\cdot)$ is the token-weighted cross-entropy loss. The difference between $\mathcal{L}_w(x \mid I)$ (predicting the instruction from the image without response) and $\mathcal{L}_w(x \mid y, I)$ (with response) quantifies the information gain provided by $y$ for reconstructing $x$.
Following the same token-weighting strategy, RIC can be expressed as:
\begin{equation}
\small
\begin{split}
    \text{RIC}(I,x,y) = \exp\Bigg(&\frac{1}{\sum_{i=1}^{N} w_i} \sum_{i=1}^{N} w_i \Big[
    -\log p(t_i \mid I, t_{<i}) \\
    &+ \log p(t_i \mid y, I, t_{<i}) \Big]\Bigg).
\end{split}
\end{equation}
Intuitively, if $y$ aids predicting instruction tokens ($p(t_i \mid y, I, t_{<i}) > p(t_i \mid I, t_{<i})$), it positively contributes to the RIC score. High scores identify samples with coherent semantics alignment. For multi-turn dialogues, we average scores across turns to capture overall consistency.

\paragraph{Data Intrinsic Consistency.}
For each sample, we define its Data Intrinsic Consistency (DIC) score by taking the minimum of VIC and RIC:
\begin{equation}
\label{eq:dic}
    \text{DIC}(I,x,y) = \min\!\left(\text{VIC}(I,x,y),\, \text{RIC}(I,x,y)\right).
\end{equation}
Eq.~\ref{eq:dic} enforces that high-quality samples must exhibit strong consistency across \emph{both dimensions}: the image must be informative for generating the response, and the response must be precisely aligned with the instruction. 
\section{Instruction Data Selection}
\label{sec:selection}

Given a visual instruction dataset $\mathcal{D}$, our goal is to select a subset $\mathcal{D}^*$ based on DIC scores and a sampling ratio $p$. The selection process is guided by three key principles:
(1) \emph{High DIC Score}: Prioritize samples with high DIC scores to maximize information consistency during training.
(2) \emph{Diversity Preservation}: Maintain semantic coverage to prevent distribution shift in the selected subset.
(3) \emph{Budget Adaptability}: Balance DIC quality and semantic diversity across sampling ratio $p$.

To this end, we propose \textbf{Data Intrinsic Consistency Selection (DICS)}, an adaptive data selection method based on DIC. Fig.~\ref{fig:pipeline} illustrates the data selection pipeline, while Algorithm~\ref{alg:dic_select} in appendix shows the algorithmic pseudocode. The selection process comprises two stages:
(1) \emph{Scoring and Ranking}: Compute the DIC score for each sample in $\mathcal{D}$ and rank them accordingly.
(2) \emph{Adaptive Sampling}: Perform an \emph{adaptive selection strategy} based on the DIC ranking and the target ratio $p$ using the Farthest Point Sampling (FPS) algorithm.
Specifically, the \emph{adaptive selection strategy} adapts to the sampling ratio $p$ as follows:
\begin{itemize}[leftmargin=*,nosep]
    \item \textbf{Low Ratio} ($p<50\%$): A candidate pool is formed consisting of the top-$2p$ fraction of the ranked samples. The FPS then selects the fraction $p$ from this pool to balance quality and diversity.
    \item \textbf{Boundary} ($p=50\%$): Top 50\% samples are directly selected.
    \item \textbf{High Ratio} ($p>50\%$): The top-$(2p-1)$ fraction forms the core set. Afterward, FPS selects an additional $(1-p)$ fraction from the remaining samples to fill the quota.
\end{itemize}
The core idea behind this selection strategy is twofold: when selecting a small proportion of data, the method prioritizes high-DIC samples while maintaining diversity; conversely, when selecting a large proportion, it removes redundant samples with low DIC scores from the pool. This strategy for large-scale selection mirrors practical filtering approaches, effectively striking a balance between selection efficiency and data quality.

\section{Experiments}
\label{sec:experiments}

\begin{table*}[t]
    \centering
    \caption{\textbf{Performance comparison of data selection methods using 25\% of the LLaVA-1.5-665K dataset.} The best and second-best results are highlighted in \textbf{bold} and \underline{underlined}, respectively.}
    \resizebox{\textwidth}{!}{
        \begin{tabular}{lc ccc ccc ccc cc c c}
            \toprule
            & & \multicolumn{3}{c}{\textbf{General Capability}} 
            & \multicolumn{3}{c}{\textbf{Knowledge \& Reasoning}} 
            & \multicolumn{3}{c}{\textbf{Document Understanding}} 
            & \multicolumn{2}{c}{\textbf{Hallucination}} 
            & \multicolumn{1}{c}{\textbf{Dialogue}} 
            & \\
            \cmidrule(lr){3-5} \cmidrule(lr){6-8} \cmidrule(lr){9-11} \cmidrule(lr){12-13} \cmidrule(lr){14-14}
            
            \textbf{Method} & \textbf{Scale} 
            & \textbf{MME} & \multicolumn{2}{c}{\textbf{MMBench}} 
            & \textbf{MMMU} & \textbf{SQA-I} & \textbf{VizWiz} 
            & \textbf{AI2D} & \textbf{DocVQA} & \textbf{InfoVQA} 
            & \textbf{POPE} & \textbf{Hallusion} 
            & \textbf{MMVet} & \textbf{Rel. (\%)} \\
            
            & & & \textbf{en} & \textbf{cn} & & & & & & & & \textbf{Bench} & & \\
            \midrule
            
            Full Dataset
            & 665K
            & 59.92 & 60.22 & 51.93 & 35.89 & 68.32 & 31.79 & 47.83 & 23.82 & 22.30 & 84.49 & 27.80 & 41.01 & 100.00 \\
            \midrule
            
            Random Selection
            & 166K
            & 59.28 & 56.27 & 51.93 & 35.56 & \textbf{66.98} & 29.53 & 45.08 & 21.62 & 21.63 & 81.58 & \underline{29.46} & 38.07 & 96.69 \\
            
            Length~\cite{zhao2024long}
            & 166K
            & 55.92 & 49.23 & 43.27 & 34.22 & 60.39 & 27.21 & 41.06 & 18.88 & 19.69 & 76.70 & \textbf{34.01} & 35.52 & 88.79 \\

            CLIP-Score~\cite{hessel2021clipscore} 
            & 166K
            & 58.52 & 58.20 & 51.93 & 34.33 & \textbf{66.98} & 30.32 & \textbf{51.59} & 22.26 & 21.34 & 83.46 & 26.24 & \textbf{42.80} & \underline{98.66} \\
            
            Perplexity~\cite{marion2023less} 
            & 166K
            & 54.64 & 43.19 & 41.95 & 33.78 & 57.91 & 29.68 & 46.31 & 21.64 & \underline{22.50} & 84.04 & 26.07 & 37.34 & 89.85 \\
            
            IFD~\cite{li2024quantity} 
            & 166K
            & \textbf{64.81} & 34.83 & 33.82 & 32.78 & 57.76 & 29.27 & 44.85 & 19.84 & 22.19 & 84.10 & 26.70 & 35.46 & 87.58 \\

            ARDS~\cite{yang2025data}
            & 166K
            & 55.81 & \textbf{60.06} & \underline{53.10} & \underline{35.89} & \textbf{66.98} & 27.21 & \textbf{51.59} & 18.97 & 21.49 & 81.29 & 28.01 & 35.28 & 96.46 \\
            
            COINCIDE~\cite{lee2024concept} 
            & 166K
            & 58.38 & 52.55 & 48.14 & 34.44 & \underline{66.73} & \underline{31.32} & \underline{51.04} & 23.05 & 22.46 & \underline{84.24} & 28.06 & 37.66 & 96.88 \\
            
            DataTailor~\cite{yu2025mastering} 
            & 166K
            & 59.75 & 56.42 & 51.01 & 34.22 & 65.59 & 31.25 & 46.05 & 22.64 & 21.42 & 80.24 & 28.92 & \underline{42.16} & 97.17 \\
            
            PRISM~\cite{bi2025prism} 
            & 166K
            & 60.03 & 56.58 & 51.55 & \textbf{36.11} & 66.63 & \textbf{31.77} & 46.70 & \underline{23.87} & 21.87 & 82.18 & 26.49 & 38.58 & 97.66 \\
            
            \midrule
            \rowcolor{cyan!15}  
            DICS (Ours)
            & 166K
            & \underline{63.39} & \underline{59.52} & \textbf{55.03} & \textbf{36.11} & 66.14 & 31.29 & 50.52 & \textbf{24.64} & \textbf{23.11} & \textbf{84.91} & 26.79 & 41.74 & \textbf{101.40} \\
            \bottomrule
        \end{tabular}
    }
    \vspace{-5mm}
    \label{tab:main}
\end{table*}

\subsection{Experimental Settings}
\label{sec:experiments:setup}

\textbf{Datasets \& Implementation Details.}
We conduct experiments on two widely adopted open-source visual instruction datasets: \textbf{LLaVA-1.5-665K} and \textbf{Vision-FLAN-186K}, utilizing the LLaVA-1.5-7B-Base backbone initialized from LLaVA-1.5-Captions-558K. To further validate the efficacy of DICS at scale, we curate \textbf{DICS-6M}, a dataset comprising 6 million high-quality instruction samples collected according to the official InternVL3 open-source strategy (details in Appendix~\ref{sec:appendix:internvl-6m}), and fine-tune the InternVL3-8B-Pretrained model on it.
Regarding training protocols, we fine-tune LLaVA-1.5 with LoRA on 8 NVIDIA A100 (40GB) GPUs, and fully fine-tune InternVL3 on 32 GPUs, both for one epoch. 
For DIC computation, the DIC score of text-only samples is fixed to 1.0, since no visual information is present.
To ensure stability during feature extraction (initiated from the final token of the last layer), we briefly warm up the base model on a 5\% random subset before scoring.
The warm-up checkpoint is used only for scoring and is then discarded; all final models are trained from the original pre-trained checkpoint.
Unless otherwise specified, 
all reported results are obtained on 25\% subsets of the respective full datasets.

\noindent \textbf{Benchmarks.} We assess the capabilities of fine-tuned models across 12 benchmarks spanning five distinct dimensions: (1) General Capability (MME~\cite{fu2025mme}, MMBench~\cite{liu2024mmbench}); (2) Knowledge and Reasoning (MMMU~\cite{yue2024mmmu}, SQA-I~\cite{lu2022learn}, VizWiz~\cite{gurari2018vizwiz}); (3) Document Understanding (AI2D~\cite{Kembhavi2016ADI}, DocVQA~\cite{Mathew2020DocVQAAD}, InfoVQA~\cite{Mathew2021InfographicVQA}); (4) Hallucination Mitigation (POPE~\cite{li2023evaluating}, HallusionBench~\cite{guan2024hallusionbench}); and (5) Open-ended Dialogue (MM-Vet~\cite{yu2023mm}). Further details are provided in the Appendix~\ref{sec:appendix:benchmark}.

\noindent \textbf{Baselines.} We conduct a comparative analysis of DICS against four categories of baseline methods: (1) Stochastic Sampling via random selection to ensure uniform data diversity; (2) Heuristic Filtering based on data length~\cite{zhao2024long} and CLIP-Score~\cite{radford2021learning}; (3) LLM-based Selectors, such as Perplexity~\cite{marion2023less} and IFD~\cite{li2024quantity}, which assess data quality via loss differentials within LLMs; and (4) MLLM-based Selectors, including ARDS~\cite{yang2025data}, COINCIDE~\cite{lee2024concept}, DataTailor~\cite{yu2025mastering}, and PRISM~\cite{bi2025prism}. All baselines follow their official implementations. See Appendix~\ref{sec:appendix:baselines} for details.

\subsection{Main Results}
\label{sec:experiments:main}

\noindent \textbf{Efficacy of DICS.} 
Table~\ref{tab:main} presents the results of various data selection methods on the LLaVA-1.5-7B model. Remarkably, DICS achieves an average improvement of 1.40\% across all 12 benchmarks compared to full-dataset training. This consistent gain across diverse task categories underscores that intrinsic data consistency outweighs mere scale. By leveraging DICS, we effectively filter out suboptimal samples characterized by low visual dependency or low instruction-response consistency. Specifically, our method attains state-of-the-art performance on key benchmarks, including MMBench-CN (55.03), MMMU (36.11), DocVQA (24.64), InfoVQA (23.11), and POPE (84.91).
In contrast, prior heuristic methods often trade generalization for peak performance on specific metrics. For instance, length-based selection maximizes HallusionBench scores by constraining response length, yet this reduces information density and reasoning depth, resulting in a significant drop in overall capability (88.79\% relative performance). DICS overcomes these limitations by evaluating intrinsic consistency among images, instructions, and responses, thereby ensuring balanced and robust performance across all dimensions.

\noindent \textbf{Performance across Sampling Ratios.}
As illustrated in Fig.~\ref{fig:teaser}, DICS consistently outperforms representative strong baselines across varying sampling ratios. Under low proportions ($p < 25\%$), DICS demonstrates a distinct advantage, validating that our consistency metrics effectively identify high-value samples. 
As $p$ increases, performance initially peaks (reaching 103.15\% at 50\%) before gradually declining and saturating. This occurs because higher sampling ratios incorporate more low-DIC samples, compromising model alignment and instruction-following capabilities. 
Notably, even after peaking, DICS maintains a lead over competitors. In contrast, PRISM exhibits marked degradation due to the inclusion of low-quality samples. These curves indicate that DICS rapidly identifies beneficial samples while delaying the introduction of noise, thereby maximizing data efficiency.
It is worth noting that the optimal sampling rate $p$ is not fixed but depends on specific metrics and datasets. Similarly, other methods exhibit comparable peaks; for instance, both PRISM and DataTailor reach their optimal performance around 75\%.

\begin{table}[t!]
\centering
\caption{\textbf{Performance comparison of data selection methods using 25\% of Vision-FLAN dataset.}}
\label{tab:visionflan}
\resizebox{0.49\textwidth}{!}{
\begin{tabular}{l ccccc c}
\toprule
\textbf{Method} &
\makecell{\textbf{General}} &
\makecell{\textbf{Know.}} &
\makecell{\textbf{Doc.}} &
\makecell{\textbf{Hall.}} &
\makecell{\textbf{Dial.}} &
\textbf{Rel.} \\
\cmidrule(lr){2-2} \cmidrule(lr){3-3} \cmidrule(lr){4-4} \cmidrule(lr){5-5} \cmidrule(lr){6-6}
\textit{(Sampling Ratio: 25\%)} & Avg. & Avg. & Avg. & Avg. & Avg. & \textbf{(\%)} \\
\midrule
Full Dataset & 51.93 & 42.07 & 29.54 & 58.21 & 35.28 & 100.00 \\
\midrule
Random Selection & 47.90 & 41.45 & 28.34 & 57.87 & 34.40 & 96.34 \\
Length~\cite{zhao2024long} & 45.27 & 38.90 & 24.55 & 57.75 & 36.10 & 91.47 \\
CLIP-Score~\cite{hessel2021clipscore} & 44.46 & 41.73 & 33.42 & 57.74 & 37.06 & 97.90 \\
Perplexity~\cite{marion2023less} & 30.38 & 38.28 & 27.93 & 56.51 & 33.35 & 83.50 \\
IFD~\cite{li2024quantity} & 44.13 & 39.63 & 28.92 & 55.95 & 36.38 & 93.11 \\
DataTailor~\cite{yu2025mastering} & 44.64 & 40.87 & 30.98 & 58.26 & 36.38 & 96.18 \\
PRISM~\cite{bi2025prism} & 49.14 & 42.26 & 30.28 & 58.54 & 32.29 & 98.48 \\
\midrule
\rowcolor{cyan!15}
DICS (Ours) & 47.95 & 42.25 & 32.53 & 58.30 & 37.06 & 99.91 \\
\bottomrule
\end{tabular}
}
\end{table}

\begin{table}[t]
    \centering
    \caption{\textbf{Generalization across architectures on LLaVA-1.5-665K.}
    Average performances are reported.}
    \resizebox{0.48\textwidth}{!}{
        \begin{tabular}{ll cc}
            \toprule
            \textbf{Model} & \textbf{Selector} & \textbf{Avg. Score} & \textbf{Rel.(\%)} \\
            \midrule

            \multirow{4}{*}{\textbf{LLaVA-1.5-7B}} 
            & Full Dataset     & 46.28 & 100.00 \\
            & Random Selection & 45.24 & 97.74 \\
            \cmidrule{2-4} 
            & DICS (LLaVA-1.5-7B) & 46.93 & 101.40 \\
            & DICS (Qwen2-VL-7B)  & 46.36 & 100.16 \\
            \midrule

            \multirow{4}{*}{\textbf{Qwen2-VL-7B}} 
            & Full Dataset     & 70.46 & 100.00 \\
            & Random Selection & 70.61 & 100.21 \\
            \cmidrule{2-4}
            & DICS (LLaVA-1.5-7B) & 71.35 & 101.25 \\
            & DICS (Qwen2-VL-7B)  & 71.10 & 100.91 \\
            \midrule
            
            \multirow{3}{*}{\textbf{LLaVA-1.5-13B}} 
            & Full Dataset     & 47.89 & 100.00 \\
            & Random Selection & 47.86 & 99.93 \\
            \cmidrule{2-4}
            & DICS (LLaVA-1.5-7B) & 48.75 & 101.79 \\
            
            \bottomrule
        \end{tabular}
    }
    \vspace{-3mm}
    \label{tab:generalization_matrix}
\end{table}

\subsection{Robustness and Scalability}
\label{sec:experiment:generalization}

\noindent \textbf{Cross-Dataset Generalization.}
To demonstrate robustness across diverse domains, we evaluate DICS on Vision-Flan-186K, which features a distribution distinct from LLaVA-1.5-665K (see Table~\ref{tab:visionflan}). Remarkably, DICS achieves the best overall relative performance: training on just a 25\% subset yields a relative performance of \textbf{99.91\%}, outperforming all baselines and closely matching full-dataset fine-tuning. These results confirm that DICS generalizes beyond specific data patterns, serving as a robust indicator for selection. Detailed results are listed in Appendix Table~\ref{tab:visionflan_appendix}.

\noindent \textbf{Cross-Architecture Generalization.}
To validate compatibility across diverse architectures, we extend evaluation to Qwen2-VL-7B under the same settings. As shown in Table~\ref{tab:generalization_matrix}, DICS-selected subsets (using Qwen2-VL as the selector) consistently outperform the full-dataset baseline, achieving a relative performance of \textbf{100.91\%}. This indicates that DICS prioritizes intrinsic data consistency independent of model structure. 
Beyond single-model generalization, we observe robust cross-architectural transferability: data selected by LLaVA-1.5-7B effectively enhances Qwen2-VL-7B (101.25\%), while conversely, subsets selected by Qwen also boost the LLaVA model. This mutual improvement confirms that DIC targets universal sample characteristics rather than model-specific patterns. 
Notably, the LLaVA-1.5-13B model achieves \textbf{101.79\%} performance using data selected by the smaller 7B variant. 
This suggests that DIC score computed by smaller models transfer to larger-scale training, allowing lightweight selectors to reduce the cost of larger data curation.

\begin{table}[t!]
\centering
\caption{\textbf{Performance comparison on the 6M large-scale dataset using InternVL3-8B.}}
\label{tab:scaling_6m}
\resizebox{0.49\textwidth}{!}{
\begin{tabular}{l c ccccc c}
\toprule
\multirow{2}{*}{\textbf{Method}} & \multirow{2}{*}{\textbf{Scale}} &
\makecell{\textbf{General}} &
\makecell{\textbf{Know.}} &
\makecell{\textbf{Doc.}} &
\makecell{\textbf{Hall.}} &
\makecell{\textbf{Dial.}} &
\textbf{Rel.} \\
\cmidrule(lr){3-3} \cmidrule(lr){4-4} \cmidrule(lr){5-5} \cmidrule(lr){6-6} \cmidrule(lr){7-7}
& & Avg. & Avg. & Avg. & Avg. & Avg. & \textbf{(\%)} \\
\midrule
\rowcolor{gray!10}
InternVL3-8B-Instruct & 21.7M & 84.29 & 78.80 & 82.94 & 67.79 & 70.10 & 109.08 \\
\midrule
Full Dataset (100\%) & 6.0M & 77.62 & 69.42 & 77.88 & 65.73 & 56.19 & 100.00 \\
\midrule
DICS (15\%) & 0.9M & 75.16 & 68.87 & 77.29 & 67.58 & 56.88 & 99.26 \\
DICS (25\%) & 1.5M & 77.92 & 71.71 & \textbf{78.94} & \textbf{68.07} & 54.50 & 101.47 \\
DICS (50\%) & 3.0M & 75.97 & 70.64 & 78.01 & 67.57 & 57.16 & 100.32 \\
DICS (75\%) & 4.5M & 79.79 & \textbf{72.94} & 78.47 & 67.11 & 52.89 & 101.86 \\
DICS (85\%) & 5.1M & \textbf{80.53} & 72.59 & 78.76 & 67.70 & \textbf{59.13} & \textbf{103.10} \\
\bottomrule
\end{tabular}
}
\end{table}
\noindent \textbf{Scalability across Model Architectures.}
To validate DICS under modern architectures and with large-scale data, we conducted experiments on \textbf{DICS-6M} , a 6M-sample corpus curated following InternVL3's open-source data collection protocol (details in Appendix~\ref{sec:appendix:internvl-6m}). We evaluated DICS by comparing our results against models fine-tuned with varying sampling ratios as well as the official InternVL3-8B-Instruct with DICS-6M. As shown in Table~\ref{tab:scaling_6m}, using the full 6M dataset as a baseline, DICS consistently outperforms across all sampling ratios. Notably, at sampling ratios exceeding 25\%, the fine-tuned model surpasses the performance of the full dataset. Specifically, at an 85\% sampling ratio, the model achieves optimal performance, outperforming the full-dataset baseline by 3.10\%. 
Remarkably, despite utilizing less than 25\% of the official data volume (5.1M vs. 21.7M), our model demonstrates performance comparable to the official InternVL3-8B-Instruct, achieving a relative score of 94.52\% (our 103.10\% vs. official 109.08\%).
These findings underscore two key insights: (1) the superior quality of DICS-6M and the significant redundancy inherent in the original visual instruction data; and (2) the robust scalability and practical utility of DICS.

\subsection{Ablation Studies and Analysis}
\label{sec:experiments:analysis}

\begin{table}[t!]
\centering
\caption{\textbf{Ablation study on DICS.}}
\label{tab:ablation_dic}
\resizebox{0.49\textwidth}{!}{
\begin{tabular}{l ccccc c}
\toprule
\textbf{Variant} &
\makecell{\textbf{General}} &
\makecell{\textbf{Know.}} &
\makecell{\textbf{Doc.}} &
\makecell{\textbf{Hall.}} &
\makecell{\textbf{Dial.}} &
\textbf{Rel.} \\
\cmidrule(lr){2-2} \cmidrule(lr){3-3} \cmidrule(lr){4-4} \cmidrule(lr){5-5} \cmidrule(lr){6-6}
\textit{(Sampling Ratio: 25\%)} &
Avg. & Avg. & Avg. & Avg. & Avg. & \textbf{(\%)} \\
\midrule

\multicolumn{7}{l}{\textit{\textbf{Baselines}}} \\
Full Dataset (100\%)  & 57.36 & \textbf{45.33} & 31.32 & 56.15 & 41.01 & 100.00 \\
FPS (Diversity Only)  & 54.10 & 43.40 & 31.86 & 55.76 & 39.08 & 97.02  \\
\midrule

\multicolumn{7}{l}{\textit{\textbf{Single Consistency Metric}}} \\
~~ DICS w/ VIC only & 59.03 & 44.86 & 32.73 & 55.82 & 39.54 & 101.02 \\
~~ DICS w/ RIC only & 53.91 & 43.92 & 31.45 & \textbf{56.98} & 40.28 & 97.60  \\
\midrule

\multicolumn{7}{l}{\textit{\textbf{Metric Fusion Strategies}}} \\
~~ Sum (VIC+RIC) & 56.43 & 43.80 & 31.50 & 55.40 & 41.56 & 98.61  \\
~~ Product (VIC$\times$RIC) & 56.24 & 44.53 & 31.72 & 56.20 & 39.17 & 98.84  \\
\midrule

\multicolumn{7}{l}{\textit{\textbf{Weighted Token Strategy}}} \\
~~ DICS w/o weighted token &  58.19 & 43.93 & \textbf{32.85} & 55.92 & 40.05 & 100.25 \\

\midrule
\rowcolor{cyan!15}
\textbf{DICS (Ours)}  & \textbf{59.31} & 44.51 & 32.76 & 55.85 & \textbf{41.74} & \textbf{101.40} \\
\bottomrule
\end{tabular}
}
\vspace{-4mm}
\end{table}

\noindent \textbf{Ablation Studies on DICS.}
To assess individual contributions of our metrics, we conducted ablation studies on a 25\% (166K) subset of LLaVA-1.5-665K (Table~\ref{tab:ablation_dic}). Pure diversity-based sampling (FPS, 97.02\%) underscores the necessity of quality-aware filtering. Among single metrics, VIC drives significant gains (101.02\%) while RIC specifically mitigates hallucinations. However, simple fusions (Sum/Product) fail ($\sim$98\%) because high scores in one metric can compensate for low scores in the other, introducing noise. In contrast, our Min-based DICS requires both dimensions to be high. Additionally, removing our weighted token strategy causes a drop from 101.40\% to 100.25\%, validating its importance. Ultimately, our proposed DICS achieves the peak relative performance (101.40\%) by combining visual alignment with logical coherence.

\begin{table}[t!]
\centering
\caption{\textbf{Robustness across selection strategies and combinations.} We evaluate different scoring metrics combined with various sampling methods at a 25\% ratio.}

\label{tab:diversity_strategy}
\resizebox{0.49\textwidth}{!}{
\begin{tabular}{l ccccc c}
\toprule
\textbf{Method \& Strategy} &
\makecell{\textbf{General}} &
\makecell{\textbf{Know.}} &
\makecell{\textbf{Doc.}} &
\makecell{\textbf{Hall.}} &
\makecell{\textbf{Dial.}} &
\textbf{Rel.} \\
\cmidrule(lr){2-2} \cmidrule(lr){3-3} \cmidrule(lr){4-4} \cmidrule(lr){5-5} \cmidrule(lr){6-6}
\textit{(Sampling Ratio: 25\%)} &
Avg. & Avg. & Avg. & Avg. & Avg. & \textbf{(\%)} \\
\midrule
Full Dataset              & 57.36 & \textbf{45.33} & 31.32 & 56.15 & 41.01 & 100.00 \\
\midrule
DataTailor (Top-K)        & 56.08 & 44.07 & 31.82 & 55.54 & 41.51 & 98.76 \\
DataTailor (Hierarchical) & 55.73 & 43.69 & 31.76 & 54.58 & \textbf{42.16} & 98.11 \\
DataTailor (FPS)          & 54.37 & 43.86 & 32.23 & 53.92 & 41.51 & 97.37 \\
PRISM (Top-K)             & 56.05 & 44.84 & 32.43 & 54.34 & 38.58 & 98.53 \\
PRISM (FPS)               & 54.95 & 43.86 & 32.06 & 55.32 & 40.23 & 97.86 \\
\midrule
DICS (Top-K)               & 57.04 & 44.78 & \textbf{33.06} & 55.85 & 41.28 & 100.40 \\
DICS (Hierarchical)        & 58.36 & 45.28 & 32.61 & \textbf{56.60} & 41.19 & 101.39 \\
\rowcolor{cyan!15}
\textbf{DICS (FPS, Ours)}  & \textbf{59.31} & 44.51 & 32.76 & 55.85 & 41.74 & \textbf{101.40} \\
\bottomrule
\end{tabular}
}
\vspace{-5mm}
\end{table}

\noindent \textbf{Ablation Study on Selection Strategies.}
To evaluate sampler impacts, we compared Top-K, Hierarchical Clustering, and Farthest Point Selection (FPS). As shown in Table~\ref{tab:diversity_strategy}, FPS yielded the best relative performance (\textbf{101.40\%}) at a sampling ratio of 25\%; therefore, we adopt it as our default. 
In particular, DICS consistently outperforms across all strategies, whereas DataTailor and PRISM fail to match the full-dataset baseline. This shows that DICS more accurately captures the intrinsic learning value, reliably identifying high-quality samples regardless of the selection method employed.

\begin{figure}[t!]
    \centering
    \includegraphics[width=1\linewidth]{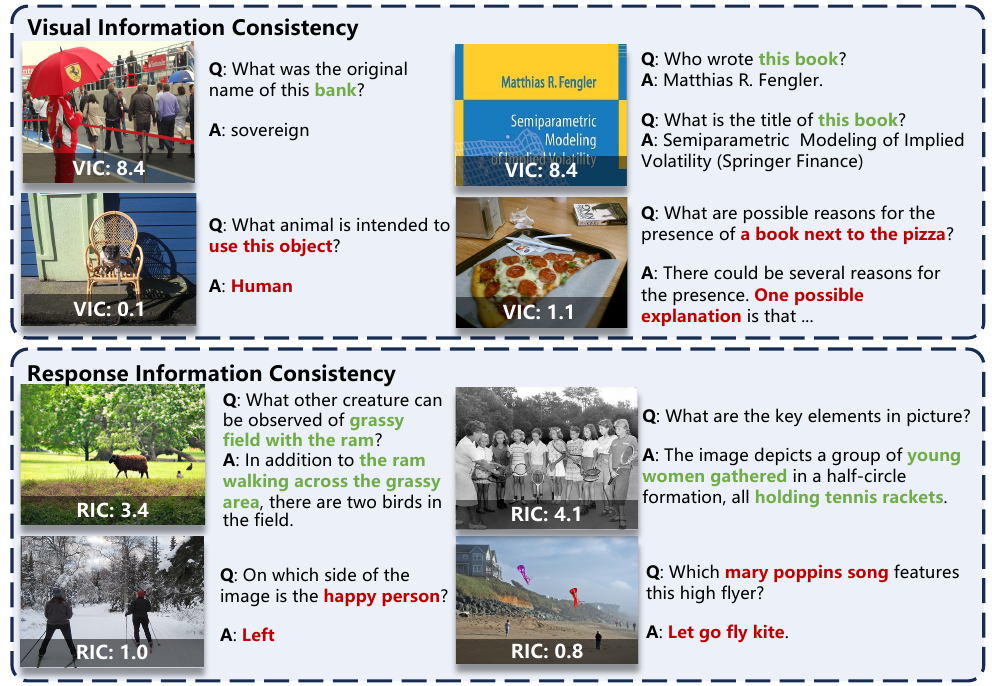}
    \caption{\textbf{Visualization of DIC}. We contrast samples with high and low VIC/RIC scores.}
    \label{fig:visualization}
\end{figure}

\noindent \textbf{Visualization of VIC and RIC.}
Fig.~\ref{fig:visualization} illustrates visual instruction samples with varying VIC and RIC scores. 
High-VIC samples are characterized by the need to rely on explicit visual cues in the image (e.g., text on logos or book covers) to answer questions; in contrast, low-VIC samples tend to be vision-irrelevant (relying solely on linguistic information) or contain hallucinated content. Complementarily, high-RIC responses exhibit strong semantic consistency and logical coherence, while low-RIC responses often suffer from disconnection and fail to capture the intended meaning. Consequently, samples scoring high in both VIC and RIC represent the ideal core instructional data.

\begin{figure}[t!]
    \centering
    \includegraphics[width=1\linewidth]{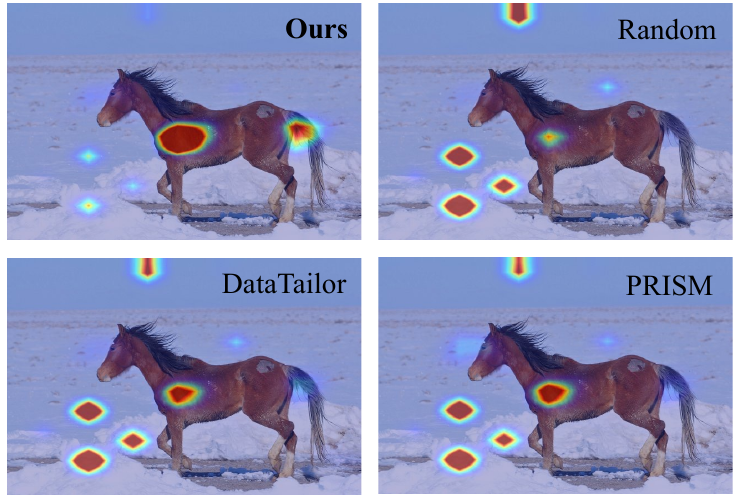}
    \caption{\textbf{Qualitative comparison on text-to-image attention maps} for the query: \textit{``What color is the horse in the image?''} (Answer: \textit{``Brown.''}). DICS exhibits more precise visual grounding.}
    \label{fig:attn_map}
\end{figure}

\noindent \textbf{Qualitative Comparison}.
Fig.~\ref{fig:attn_map} visualizes text-to-image attention maps (response tokens to visual patches) for different models. Compared to other methods, the DICS-trained model exhibits a more focused attention distribution that precisely localizes the queried object (e.g., the horse's body), indicating superior visual grounding. More details can be seen in Appendix~\ref{sec:appendix:attention_analysis}.

\noindent \textbf{Computational Efficiency.}
DICS is highly efficient, requiring only inference-time forward passes without gradient updates. On the LLaVA-1.5-665K dataset using 8 A100 GPUs, data selection took $\sim$12 hours; selecting a 25\% subset reduced total training time to 14 hours (a 1-hour saving vs. full-dataset training). In the DICS-6M experiment with 32 A100 GPUs, selection and training were completed in 85 hours, yielding a 5-hour savings over the baseline while achieving a 3.10\% performance gain. Since DIC scores are computed per sample, computational cost scales linearly with data volume. This characteristic, combined with its inherent parallelizability, makes DICS increasingly advantageous for large-scale datasets. Detailed cost breakdowns are provided in Appendix~\ref{sec:appendix:compute_cost}.
\section{Conclusion}
\label{sec:conclusion}

This paper introduces Data Intrinsic Consistency (DIC), a self-scoring metric quantifying sample-level inter-component consistency. DIC comprises two modules: \emph{Visual Information Consistency (VIC)}, evaluating visual-instruction alignment, and \emph{Response Information Consistency (RIC)}, assessing response coherence relative to the instruction. Leveraging DIC, we propose Data Intrinsic Consistency Selection (DICS), an adaptive selection method balancing intra-sample consistency with distributional diversity across varying data budgets. Extensive experiments demonstrate that DICS consistently outperforms SoTA baselines across diverse dataset scales and model architectures, validating its robustness and scalability.

\noindent \textbf{Limitations and Future Work.} 
A primary limitation is that our consistency evaluation currently focuses on image-text data, excluding modalities like video or audio; we plan to extend this framework to broader multimodal contexts in future work. We identify two promising directions: (1) developing an adaptive strategy to automatically determine optimal subset size and composition, and (2) shifting from discarding low-DIC samples to actively repairing them to enhance model performance. 

\noindent \textbf{Ethical Statement.}
In this work, we utilize publicly available visual instruction tuning datasets for visual instruction selection to facilitate reproducibility. Concurrently, following the data curation protocols outlined in InternVL, we have compiled an additional set of data, which will be publicly released in the near future (see Appendix~\ref{sec:datasets}). However, certain samples within these datasets contain erroneous answers regarding the images that lack a clear alignment with the provided instructions and responses. Fine-tuning Large Vision-Language Models (LVLMs) on such flawed data can cause the models to generate incorrect visual interpretations or suffer from hallucinations. To address this issue, our proposed method aims to effectively choose high-quality samples, thereby improving the efficacy of instruction tuning. Therefore, our work does not introduce any unforeseen ethical concerns.

\section{Acknowledgements}
This work was supported by the 2035 Innovation Mission Project  of CASIA (No.E4J10102).

\bibliography{custom}

\newpage
\appendix
\clearpage
\section{Implementation Details}

\subsection{Token Weighting Strategy}
\label{sec:appendix:token_weight}

To enhance semantic sensitivity in both VIC (Sec.~\ref{sec:dic:vic}) and RIC (Sec.~\ref{sec:dic:ric}), we employ a heuristic-based token weighting strategy. Specifically, we assign a full weight of $1.0$ to content words to prioritize semantics, and a reduced weight of $0.1$ to functional words to prevent frequent but grammatically necessary tokens from dominating the calculation. 

We utilize the spaCy library~\cite{honnibal2020spacy} to automatically identify the Part-of-Speech (POS) tag for each token and assign these weights accordingly. Formally, let $\text{POS}(t_i)$ denote the part-of-speech tag of token $t_i$. We define the weight $w_i$ as:
\begin{equation}
w_i = 
\begin{cases}
1.0, & \text{if } \text{POS}(t_i) \in \mathcal{C}, \\
0.1, & \text{if } \text{POS}(t_i) \in \mathcal{F}, \\
1.0, & \text{otherwise}.
\end{cases}
\end{equation}
where $\mathcal{C}$ and $\mathcal{F}$ denote content words and functional words, respectively.

\paragraph{Content Words ($\mathcal{C}$).} These are semantically meaningful tokens that carry the core information of a sentence. To prioritize their impact, we assign full weight ($w_i = 1.0$) to the following POS categories:
\begin{itemize}
    \item \textbf{Nouns} (NN, NNS, NNP, NNPS): such as ``dog'', ``image'', and ``building'';
    \item \textbf{Verbs} (VB, VBD, VBG, VBN, VBP, VBZ): such as ``describe'', ``show'', and ``contains'';
    \item \textbf{Adjectives} (JJ, JJR, JJS): such as ``red'', ``large'', and ``beautiful'';
    \item \textbf{Adverbs} (RB, RBR, RBS): such as ``clearly'', ``very'', and ``quickly'';
    \item \textbf{Numbers} (CD): such as ``three'' and ``2024''.
\end{itemize}

\paragraph{Functional Words ($\mathcal{F}$).} These tokens primarily serve grammatical purposes and contribute less to the semantic content. To prevent them from dominating the metric, we assign reduced weight ($w_i = 0.1$) to:
\begin{itemize}
    \item \textbf{Determiners} (DT): such as ``the'', ``a'', ``an'', and ``this'';
    \item \textbf{Prepositions} (IN): such as ``in'', ``on'', ``at'', and ``of'';
    \item \textbf{Conjunctions} (CC): such as ``and'', ``but'', and ``or'';
    \item \textbf{Pronouns} (PRP, PRP\$, WP, WP\$): such as ``it'', ``they'', and ``which'';
    \item \textbf{Auxiliary Verbs} (MD): such as ``can'', ``will'', and ``should'';
    \item \textbf{Punctuation} : such as ``.'', ``,'', ``:'', etc;
    \item \textbf{Particles} (RP, TO): such as ``to'' and ``up''.
\end{itemize}
This rigorous weighting scheme ensures that our proposed metrics effectively capture semantically significant elements while filtering out noise from informationally sparse components.

\subsection{Implementation Details of VIC}
\label{sec:appendix:vic_impl}

\noindent \textbf{Image Blocking.} To compute the loss $\mathcal{L}_w(y \mid x)$ without visual information, we need to remove the influence of the image while preserving the input sequence structure. Our default approach is to \textbf{mask out the image tokens} from the input sequence, preventing any visual information from being attended to during response generation. This ensures a clean ablation of visual information without introducing distributional artifacts.
We also explored an alternative approach: \textbf{replacing the original image with an unrelated image} randomly sampled from the dataset. This substitution strategy disrupts the semantic alignment between the image and the instruction-response pair while maintaining the natural distribution of visual features. Empirically, both approaches yield comparable results, suggesting that the VIC metric is robust to the specific implementation of visual information removal.

\noindent \textbf{Text-only Samples.} For samples without images (i.e., pure text instruction data), the VIC score is $1.0$ according to Eq.~\ref{eq:vic}, indicating neutral visual information gain.

\noindent \textbf{Multi-image Samples.} When multiple images are present, all images are processed together as the visual context $I$. The VIC metric measures the collective information gain from all images.

\subsection{Implementation Details of RIC}
\label{sec:appendix:ric_impl}

As discussed in Sec.~\ref{sec:dic:ric}, the inverse formulation is essential to explicitly penalize generic, low-information responses (e.g., ``This is an image.''), which might achieve trivially low loss in standard forward prediction despite lacking specific relevance to the instruction. To compute the RIC score, we guide the model to generate the original instruction $x$ and measure the cross-entropy loss on its output tokens under the following two prompt templates:
\begin{itemize}
    \item \textbf{Without Response} (for computing $\mathcal{L}_w(x \mid I)$): \texttt{``Based on the image, what is the likely question?''}
    \item \textbf{With Response} (for computing $\mathcal{L}_w(x \mid y, I)$): \texttt{``Based on the image and answer, what is the likely question? Answer: \{y\}''}
\end{itemize}
This design ensures that only responses providing instruction-specific semantic cues attain high RIC scores.

\subsection{Implementation Details of DICS}
\label{sec:appendix:dics}

In this section, we provide the detailed algorithmic procedures for our proposed Data Intrinsic Consistency Selection (DICS) method, which synergistically combines quality-based ranking with diversity-aware sampling.

\noindent \textbf{Data Intrinsic Consistency Selection (DICS).}
As outlined in Section~\ref{sec:selection}, DICS aims to select a high-quality and diverse subset from the original dataset based on DIC scores. The complete procedure is detailed in Algorithm~\ref{alg:dic_select}. First, we compute the DIC score for each sample, defined as the minimum of its VIC and RIC scores, and sort the dataset in descending order. Then, we apply an adaptive selection strategy based on the target selection ratio $p$. For a low ratio ($p < 0.5$), we construct a candidate pool of size $2p \cdot |\mathcal{D}|$ containing the top-ranked samples, and perform diversity-aware sampling to select the final subset. For a high ratio ($p > 0.5$), we directly retain the top $(2p - 1) \cdot |\mathcal{D}|$ samples as a core set, and apply diversity-aware sampling on the remaining samples to rescue the most diverse ones to meet the budget. At the boundary ($p = 0.5$), we simply select the top 50\% samples.

\noindent \textbf{Farthest Point Sampling (FPS).}
To ensure diversity within the selected subsets, DICS utilizes Farthest Point Sampling (FPS) as a core subroutine, detailed in Algorithm~\ref{alg:fps}. FPS is a greedy algorithm that iteratively selects the sample most distant from the currently selected set, thereby maximizing the coverage of the embedding space. By applying FPS on dynamically narrowed candidate pools (as dictated by Algorithm~\ref{alg:dic_select}) rather than the full dataset, our approach effectively balances data quality (prioritizing high DIC scores) and semantic diversity. Furthermore, this piecewise sampling strategy significantly reduces the computational overhead compared to running FPS on the entire dataset.

\begin{algorithm}[t]
\caption{Data Intrinsic Consistency Selection (DICS).}
\label{alg:dic_select}
\begin{algorithmic}[1]
\Require Dataset $\mathcal{D}$, selection ratio $p \in (0, 1]$
\Ensure Selected subset $\mathcal{D}^*$ with $|\mathcal{D}^*| = p \cdot |\mathcal{D}|$

\State Compute $s(d) = \min(\text{VIC}(d), \text{RIC}(d))$ for each $d \in \mathcal{D}$
\State Sort $\mathcal{D}$ by $s(d)$ descending $\rightarrow \mathcal{D}_{\text{sorted}}$
\State $n \gets |\mathcal{D}|$, \quad $k \gets p \cdot n$

\If{$p = 0.5$}
    \State $\mathcal{D}^* \gets$ top-$k$ from $\mathcal{D}_{\text{sorted}}$
\ElsIf{$p < 0.5$}
    \State $\tilde{\mathcal{D}} \gets$ top-$2k$ from $\mathcal{D}_{\text{sorted}}$
    \State $\mathcal{D}^* \gets \text{FPS}(\tilde{\mathcal{D}}, k)$
\Else
    \State $\mathcal{D}_{\text{core}} \gets$ top-$(2k-n)$ from $\mathcal{D}_{\text{sorted}}$
    \State $\tilde{\mathcal{D}} \gets \mathcal{D}_{\text{sorted}} \setminus \mathcal{D}_{\text{core}}$
    \State $k_r \gets k - |\mathcal{D}_{\text{core}}|$
    \State $\mathcal{D}_{\text{rescue}} \gets \text{FPS}(\tilde{\mathcal{D}}, k_r)$
    \State $\mathcal{D}^* \gets \mathcal{D}_{\text{core}} \cup \mathcal{D}_{\text{rescue}}$
\EndIf

\State \Return $\mathcal{D}^*$
\end{algorithmic}
\end{algorithm}

\begin{algorithm}[h]
\caption{Farthest Point Sampling (FPS)}
\label{alg:fps}
\begin{algorithmic}[1]
\Require Candidate set $\tilde{\mathcal{D}}$ with embeddings $\{\mathbf{e}_i\}_{i=1}^{M}$, target size $k$
\Ensure Selected indices $\mathcal{S}$ with $|\mathcal{S}| = k$

\State Initialize $\mathcal{S} \gets \{\arg\max_i \|\mathbf{e}_i\|\}$ 
\State Initialize distance array $d_i \gets \|\mathbf{e}_i - \mathbf{e}_{\mathcal{S}[0]}\|_2$ for all $i$

\For{$j = 2$ to $k$}
    \State $i^* \gets \arg\max_{i \notin \mathcal{S}} d_i$ 
    \State $\mathcal{S} \gets \mathcal{S} \cup \{i^*\}$
    \For{$i \notin \mathcal{S}$}
        \State $d_i \gets \min(d_i, \|\mathbf{e}_i - \mathbf{e}_{i^*}\|_2)$
    \EndFor
\EndFor

\State \Return $\mathcal{S}$
\end{algorithmic}
\end{algorithm}

\noindent \textbf{Complexity Analysis.}
Let $M = |\tilde{\mathcal{D}}|$ denote the candidate pool size, $k$ denote the target selection size, and $d$ denote the embedding dimension. The time complexity of FPS is $O(k \cdot M \cdot d)$: each of the $k$ iterations requires computing distances to all $M$ candidates, with each distance computation taking $O(d)$ time. The space complexity is $O(M \cdot d)$ for storing embeddings plus $O(M)$ for the distance array.

In practice, we leverage GPU-accelerated batched distance computation to significantly reduce the wall-clock time. For a dataset of one million samples with 4096-dimensional embeddings, selecting 100K samples typically completes in minutes on a single A100 GPU.

\section{Experiment Details}
\subsection{Datasets} 
\label{sec:datasets}
To strictly validate the effectiveness, robustness, and scalability of our proposed DICS method, we utilize three datasets spanning different data scales, distributions, and task types. All data collection complies with license requirements.

\noindent \textbf{LLaVA-1.5-665K}~\cite{liu2023visual} is a comprehensive large-scale visual instruction tuning dataset. It aggregates 665,298 samples from diverse sources to enhance the model's multi-modal capabilities. Specifically, it consists of: (1) 158K GPT-generated multimodal instruction-following data from LLaVA-1.0-Instruct, (2) academic VQA datasets including VQAv2, GQA, and OK-VQA for visual reasoning, (3) OCR-related data from OCRVQA and TextCaps to bolster text-recognition abilities, and (4) 40K pure-language conversations from ShareGPT to maintain linguistic proficiency. 

\noindent \textbf{Vision-FLAN-186K}~\cite{xu2024vision} is a large-scale, multi-task visual instruction dataset specifically designed to address the limited task diversity in existing benchmarks. Unlike LLaVA-1.5-665K, which predominantly relies on conversational data generated by LLMs, Vision-FLAN is constructed by integrating 191 diverse vision-language tasks from various academic sources. It comprises 186,060 instances that cover a wide range of capabilities, including fine-grained recognition, spatial reasoning, and complex scene understanding.

\begin{figure}[t!]
    \centering
    \includegraphics[width=1\linewidth]{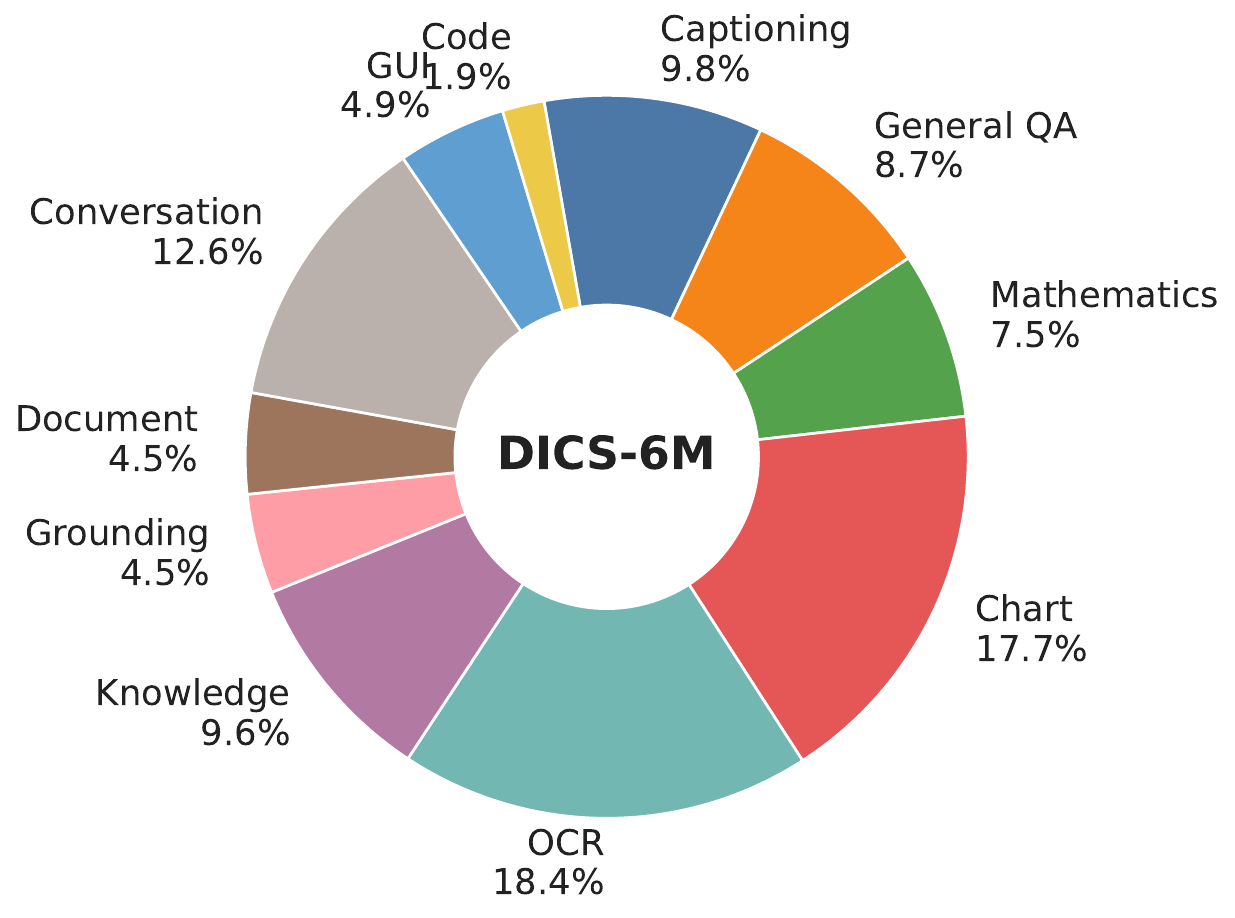}
    \caption{\textbf{Task distribution of the DICS-6M instruction dataset.} The collected corpus encompasses diverse tasks with a balanced distribution, facilitating a robust evaluation of large-scale data selection.}
    \label{fig:intern_6m_pie}
    \vspace{-3mm}
\end{figure}

\label{sec:appendix:internvl-6m}
\noindent \textbf{DICS-6M} is a large-scale visual instruction tuning corpus used to validate whether DICS remains effective in million-level data selection scenarios. Following the data collection protocol of InternVL~\cite{chen2024internvl, zhu2025internvl3}, we collect 6M public high-quality vision-language SFT samples from widely used open-source datasets. As detailed in Table~\ref{tab:intern_6m_dataset}, the corpus covers a broad range of capabilities, including general vision, document understanding, mathematical reasoning, OCR, chart understanding, GUI, etc. The category distribution in Figure~\ref{fig:intern_6m_pie} further shows that the corpus is not dominated by a single narrow source, which supports a rigorous and reliable evaluation of large-scale data selection.

\begin{table*}[t!]
\centering
\providecommand{\dataset}[1]{#1}
\caption{\textbf{Composition of the DICS-6M instruction dataset.}}
\label{tab:intern_6m_dataset}
\scriptsize
\begin{tabular}{m{0.14\textwidth}m{0.80\textwidth}}
\toprule
\textbf{Task} & \textbf{Datasets} \\
\midrule
\textbf{Captioning} & \dataset{TextCap}, \dataset{LVIS-Instruct4V}, \dataset{ShareGPT4V}, \dataset{ShareGPT4o}, \dataset{InternVL}, \dataset{COCO}, \dataset{NewYorker}, \dataset{DenseFusion}, \dataset{OCR}, \dataset{chinesememe}, \dataset{GPT-4o}, \dataset{Memotion}, \dataset{Edraw}, \dataset{Clothes Caption}, \dataset{AIG Share} \\
\rowcolor{gray!10}
\textbf{General QA} & \dataset{GQA}, \dataset{OKVQA}, \dataset{VQAv2}, \dataset{Visual7W}, \dataset{VSR}, \dataset{Objects365}, \dataset{IconQA}, \dataset{VQAonBD}, \dataset{Hateful Memes}, \dataset{FSC147}, \dataset{K12 Printing}, \dataset{CLEVR}, \dataset{LNQA}, \dataset{RobuT}, \dataset{COCO}, \dataset{Vision-FLAN}, \dataset{GPT-4o}, \dataset{Image Textualization}, \dataset{RAVEN}, \dataset{COCO-QA}, \dataset{VizWiz}, \dataset{Douban}, \dataset{Vision Oriented}, \dataset{CoSyn}, \dataset{LLRV}, \dataset{IDK}, \dataset{Ctrip}, \dataset{SketchyVQA}, \dataset{OODVQA}, \dataset{FinQA}, \dataset{VQA-AS}, \dataset{YesBut}, \dataset{SPARK}, \dataset{SVRD}, \dataset{Indoor QA}, \dataset{AIG Share}, \dataset{study\_com}, \dataset{gpt4v} \\
\textbf{Mathematics} & \dataset{MAVIS}, \dataset{MapQA}, \dataset{GeoQA}, \dataset{Geometry3K}, \dataset{UniGeo}, \dataset{GEOS}, \dataset{CLEVR-Math}, \dataset{GeomVerse}, \dataset{LaTeX Formulas}, \dataset{SynthFormulaNet}, \dataset{MathWriting}, \dataset{TallyQA}, \dataset{Geo170K}, \dataset{CoSyn}, \dataset{CLEVR}, \dataset{Gaokao}, \dataset{InterGPS}, \dataset{PicMath}, \dataset{Geo3K} \\
\rowcolor{gray!10}
\textbf{Chart} & \dataset{ChartQA}, \dataset{PlotQA}, \dataset{FigureQA}, \dataset{LRV-Instruction}, \dataset{ArxivQA}, \dataset{MMC-Inst}, \dataset{TabMWP}, \dataset{DVQA}, \dataset{UniChart}, \dataset{SimChart9K}, \dataset{Chart2Text}, \dataset{FinTabNet}, \dataset{SciTSR}, \dataset{TinyChart}, \dataset{SynthChartNet}, \dataset{ECD}, \dataset{MMTab}, \dataset{Datik}, \dataset{CoSyn}, \dataset{SBT Chart}, \dataset{Wired Table}, \dataset{VisText}, \dataset{HiTab}, \dataset{Oroikon}, \dataset{TAT-QA}, \dataset{Charts2500} \\
\textbf{OCR} & \dataset{OCRVQA}, \dataset{InfoVQA}, \dataset{TextVQA}, \dataset{ArT}, \dataset{HME100K}, \dataset{COCO-Text}, \dataset{CTW}, \dataset{LSVT}, \dataset{RCTW-17}, \dataset{VCR}, \dataset{EST-VQA}, \dataset{ST-VQA}, \dataset{EATEN}, \dataset{LLaVAR}, \dataset{CASIA}, \dataset{Chinese-OCR}, \dataset{IAM}, \dataset{NAF}, \dataset{POIE}, \dataset{ReCTs}, \dataset{MTWI}, \dataset{TextOCR}, \dataset{SROIE}, \dataset{Synthetic ArXiv OCR}, \dataset{Synthetic Infographic2Markdown}, \dataset{Synthetic OCR}, \dataset{WIT}, \dataset{SynthDog}, \dataset{TAL-OCR}, \dataset{K12 Printing}, \dataset{olmOCR}, \dataset{OCR}, \dataset{Captcha}, \dataset{LaTeX-OCR}, \dataset{CyrillicHandwriting}, \dataset{LaTeX Handwritten}, \dataset{ICDAR 2019}, \dataset{HW-SQuAD}, \dataset{Rendered Text}, \dataset{Chrome Writing}, \dataset{AIG Share}, \dataset{LaTeX QA}, \dataset{Invoices \& Receipts}, \dataset{MTVQA}, \dataset{imgur5k}, \dataset{ORAND-CAR}, \dataset{IIIT5K}, \dataset{Handwriting Forms}, \dataset{Thai OCR}, \dataset{MapText} \\
\rowcolor{gray!10}
\textbf{Knowledge} & \dataset{KVQA}, \dataset{A-OKVQA}, \dataset{ViQuAE}, \dataset{iNaturalist2018}, \dataset{MovieNet}, \dataset{ART500K}, \dataset{KonIQ-10K}, \dataset{TQA}, \dataset{ChemVLM}, \dataset{ScienceQA}, \dataset{AI2D}, \dataset{VQA-RAD}, \dataset{Wikipedia}, \dataset{Google Landmarks}, \dataset{Chinese Culture}, \dataset{Face Emotion}, \dataset{Diagram}, \dataset{CoSyn}, \dataset{Gaokao}, \dataset{VisualMRC}, \dataset{BlockDiagram} \\
\textbf{Grounding} & \dataset{COCO-ReM}, \dataset{V3Det}, \dataset{All-Seeing-V2}, \dataset{TolokaVQA}, \dataset{GPT4Gen-RD-BoxCoT}, \dataset{VisualWebInstruct}, \dataset{Downstream Grounding}, \dataset{Localized Narratives}, \dataset{OpenApp}, \dataset{AIG Share}, \dataset{SpatialSense} \\
\rowcolor{gray!10}
\textbf{Document} & \dataset{Sujet-Finance-QA-Vision}, \dataset{UReader}, \dataset{allenai\_pixmo\_docs}, \dataset{ArXiv Figures}, \dataset{CoSyn}, \dataset{InfographicVQA}, \dataset{Bentham}, \dataset{PDF-VQA}, \dataset{OmniDocBench}, \dataset{Layout Extract}, \dataset{Schedule Extract}, \dataset{FUNSD} \\
\textbf{Conversation} & \dataset{ALLaVA}, \dataset{SVIT}, \dataset{Cambrian}, \dataset{LAION-GPT4V}, \dataset{WildVision}, \dataset{Viet-ShareGPT4o}, \dataset{llava\_instruct}, \dataset{vflan}, \dataset{MMEvol}, \dataset{llava\_cot\_100k}, \dataset{llava\_wild}, \dataset{RLAIF-V}, \dataset{Visual Chat}, \dataset{GPT-4o}, \dataset{AIG Share} \\
\rowcolor{gray!10}
\textbf{GUI} & \dataset{Screen2Words}, \dataset{WebSight}, \dataset{Widget-Caption}, \dataset{RICOSCA}, \dataset{SeeClick}, \dataset{ScreenQA}, \dataset{AMEX}, \dataset{Android UI}, \dataset{UIBert}, \dataset{WaveUI}, \dataset{RootsAutomation}, \dataset{AlfredPLPL}, \dataset{Phone Action}, \dataset{Web Collected}, \dataset{AlfWorldGPT}, \dataset{Home Screen}, \dataset{Taobao App}, \dataset{Airplane App}, \dataset{WeChat App} \\
\textbf{Code} & \dataset{SynthCodeNet}, \dataset{Datikz}, \dataset{Drawing2HTML} \\
\bottomrule
\end{tabular}
\vspace{-2mm}
\end{table*}

\subsection{Benchmarks}
\label{sec:appendix:benchmark}
We conduct a comprehensive evaluation to assess the multi-dimensional capabilities of the trained models. Following the standard evaluation protocols of LLaVA~\cite{liu2023visual}, we extend the benchmark suite to include a broader range of domains. To the best of our knowledge, our evaluation covers the most extensive set of benchmarks among current multi-modal data selection studies, categorized into five distinct domains:
\begin{itemize}
    \item \textbf{General Capability:} We use \textbf{MME}~\cite{fu2025mme} and \textbf{MMBench}~\cite{liu2024mmbench} (both English and Chinese versions). MME evaluates perception and cognition across 14 subtasks, while MMBench employs a circular evaluation strategy to robustly test model capabilities. We normalize the scores on MME for computing average performance.
    
    \item \textbf{Knowledge \& Reasoning:} We employ \textbf{MMMU}~\cite{yue2024mmmu} for college-level multidisciplinary reasoning requiring domain-specific knowledge. \textbf{SQA-I (ScienceQA-Image)}~\cite{lu2022learn} assesses scientific reasoning with multi-modal contexts, and \textbf{VizWiz}~\cite{gurari2018vizwiz} tests the model's ability to answer visual questions originating from blind users, reflecting real-world applicability.
    
    \item \textbf{Document Understanding:} Recognizing the importance of fine-grained text perception, we include \textbf{AI2D}~\cite{Kembhavi2016ADI} for diagram understanding, \textbf{DocVQA}~\cite{Mathew2020DocVQAAD} for document comprehension, and \textbf{InfoVQA}~\cite{Mathew2021InfographicVQA} for infographic reasoning.
    
    \item \textbf{Hallucination:} We focus on model reliability using \textbf{POPE}~\cite{li2023evaluating}, which evaluates object existence hallucination via polling, and \textbf{HallusionBench}~\cite{guan2024hallusionbench}, which focuses on visual illusions and reasoning consistency.
    
    \item \textbf{Open-ended Dialogue:} We utilize \textbf{MM-Vet}~\cite{yu2023mm} to evaluate integrated capabilities in solving complex, open-ended problems that require reasoning, recognition, and generation.
\end{itemize}

\noindent \textbf{Evaluation Protocol.} 
For all evaluations, we utilize VLMEvalKit~\cite{duan2024vlmevalkit}, a standardized and reproducible evaluation framework. To ensure deterministic and fair comparisons, we set the generation temperature to $0.0$ for all benchmarks.

For benchmarks requiring an LLM-based judge (e.g., MM-Vet and open-ended responses in MM-Bench variants), we employ Qwen3-32B as the judge model. This choice ensures a high correlation with human judgment while maintaining evaluation efficiency. 

\subsection{Implementation Details of Baselines}
\label{sec:appendix:baselines}
We compare our framework against a comprehensive set of data selection strategies, categorized into four groups: 
(1) \textbf{Stochastic Sampling}: \textit{Random Selection} serves as the fundamental baseline, uniformly sampling a fixed proportion of data from the instruction set. 
(2) \textbf{Heuristic Filtering}: This category includes \textit{Length-based filtering}, which prioritizes samples based on sequence length, and \textit{CLIP-Score}~\cite{radford2021learning}, which selects data based on coarse-grained image-text alignment, where we utilize the clip-vit-large-patch14 model to retain samples with the highest image-question similarity.
(3) \textbf{LLM Selectors}: We evaluate strategies relying on intrinsic language model properties, including \textit{Perplexity}~\cite{marion2023less}, which estimates sample probability to filter out abnormal or high-noise data (employing LLaVA-1.5-7B for mid-range selection following prior work), and \textit{IFD}~\cite{li2024quantity}, which assesses the instruction following difficulty by measuring the loss differential between instruction-conditioned and unconditioned responses.
(4) \textbf{MLLM Selectors}: We compare against advanced strategies specifically designed for multimodal instruction tuning, including \textit{COINCIDE}~\cite{lee2024concept}, \textit{DataTailor}~\cite{yu2025mastering} and \textit{PRISM}~\cite{bi2025prism}. We strictly follow the official implementations and default settings for all advanced baselines.

\subsection{Hyperparameter Settings}
Table~\ref{tab:hyperparameters} summarizes the detailed hyperparameter settings and implementation specifics of our DICS framework. Unless otherwise specified, we adopt these default configurations across all experiments to ensure consistency. Notably, the POS-tag weighting scheme is designed to emphasize semantic content while suppressing syntactic noise, and the adaptive selection strategies ensure optimal data utilization across varying sampling ratios.

\begin{table*}[t]
    \centering
    \footnotesize 
    \renewcommand{\arraystretch}{0.85} 
    \caption{Detailed hyperparameter settings and implementation specifics of DICS.}
    \label{tab:hyperparameters}

    \begin{tabularx}{\textwidth}{
        @{\hspace{0pt}} 
        p{0.17\textwidth} 
        X 
        @{\hspace{0.2cm}} 
        p{0.17\textwidth} 
        X 
        @{\hspace{0pt}}
    }
        \toprule
        \textbf{Parameter} & \textbf{Configuration} & \textbf{Parameter} & \textbf{Configuration} \\
        \midrule
        POS weights (content) & 1.0 (nouns, verbs, adj., adv., numbers) 
        & FPS feature source & Final-token hidden rep. from the last layer \\
        
        POS weights (functional) & 0.1 (det., prep., conj., pron., aux., punct., particles) 
        & Low-ratio strategy ($\alpha < 0.5$) & Build top-$k$ pool by DIC, select remaining via FPS \\
        
        POS weights (others) & 1.0 (default) 
        & Boundary strategy ($\alpha = 0.5$) & Directly select top 50\% samples by DIC ranking \\
        
        RIC prompt (w/o resp.) & ``Based on the image, what is the likely question?'' 
        & High-ratio strategy ($\alpha > 0.5$) & Keep top-$k$ as core set, select additional via FPS \\
        
        RIC prompt (w/ resp.) & ``Based on image and answer, what is the likely question? Ans: \{y\}'' 
        & Scoring warm-up subset & 5\% random subset \\
        \bottomrule
    \end{tabularx}
\end{table*}

\section{Experiments and Analysis}
In this section, we provide a comprehensive analysis to complement the experimental results presented in Section~\ref{sec:experiments}. Specifically, we delineate performance across individual benchmarks to ensure full transparency. Furthermore, we present extended comparative experiments and ablation studies to corroborate the robustness and effectiveness of our proposed DICS framework.

\subsection{More Results}

\begin{table*}[t]
    \centering
    \caption{\textbf{Detailed performance comparison on Vision-FLAN dataset.} All methods are evaluated by selecting a 25\% subset of the training data.}
    \resizebox{\textwidth}{!}{
        \begin{tabular}{l ccc ccc ccc cc c c}
            \toprule
            & \multicolumn{3}{c}{\textbf{General Capability}} 
            & \multicolumn{3}{c}{\textbf{Knowledge \& Reasoning}} 
            & \multicolumn{3}{c}{\textbf{Document Understanding}} 
            & \multicolumn{2}{c}{\textbf{Hallucination}} 
            & \multicolumn{1}{c}{\textbf{Dialogue}} 
            & \\
            \cmidrule(lr){2-4} \cmidrule(lr){5-7} \cmidrule(lr){8-10} \cmidrule(lr){11-12} \cmidrule(lr){13-13}
            
            \textbf{Method} 
            & \textbf{MME} & \multicolumn{2}{c}{\textbf{MMBench}} 
            & \textbf{MMMU} & \textbf{SQA-I} & \textbf{VizWiz} 
            & \textbf{AI2D} & \textbf{DocVQA} & \textbf{InfoVQA} 
            & \textbf{POPE} & \textbf{Hallusion} 
            & \textbf{MMVet} & \textbf{Rel. (\%)} \\
            
            & & \textbf{en} & \textbf{cn} & & & & & & & & \textbf{Bench} & & \\
            \midrule
            
            Full Dataset
            & 53.93 & 53.79 & 48.07 & 35.78 & 64.45 & 25.99 & 51.20 & 20.61 & 16.82 & 83.85 & 32.57 & 35.28 & 100.00 \\
            \midrule
            
            Random Selection
            & \textbf{53.45} & 47.29 & 42.96 & \underline{35.11} & 63.91 & 25.34 & 44.85 & 21.73 & 18.43 & 80.85 & \textbf{34.89} & 34.40 & 96.34 \\

            Length~\cite{zhao2024long}
            & \underline{53.07} & 43.81 & 38.93 & 34.22 & 58.50 & 23.99 & 39.54 & 16.22 & 17.90 & 81.93 & 33.57 & 36.10 & 91.47 \\

            CLIP-Score~\cite{hessel2021clipscore}
            & 51.58 & 42.96 & 38.85 & 33.78 & \textbf{65.94} & 25.46 & \textbf{52.04} & 25.11 & \textbf{23.10} & 82.56 & 32.91 & \textbf{37.06} & 97.90 \\

            Perplexity~\cite{marion2023less} 
            & 39.59 & 29.02 & 22.52 & 30.22 & 57.91 & \underline{26.72} & 35.10 & \underline{26.13} & \underline{22.57} & 79.54 & 33.47 & 33.35 & 83.50 \\
            
            IFD~\cite{li2024quantity} 
            & 51.97 & 40.09 & 40.33 & 33.56 & \underline{65.15} & 20.18 & 49.74 & 16.08 & 20.95 & 82.26 & 29.63 & \underline{36.38} & 93.11 \\
            
            DataTailor~\cite{yu2025mastering} 
            & 52.98 & 42.41 & 38.54 & 31.67 & 65.10 & 25.83 & 46.99 & 25.29 & 20.65 & 81.91 & \underline{34.60} & \underline{36.38} & 96.18 \\

            PRISM~\cite{bi2025prism} 
            & 52.85 & \textbf{48.84} & \textbf{45.74} & \textbf{35.89} & \underline{65.15} & 25.73 & \underline{50.55} & 22.14 & 18.14 & \underline{83.04} & 34.04 & 32.29 & \underline{98.48} \\
            
            \midrule
            \rowcolor{cyan!15}  
            DICS (Ours) 
            & 50.67 & \underline{48.22} & \underline{44.97} & 34.78 & 65.05 & \textbf{26.93} & \underline{50.55} & \textbf{26.63} & 20.41 & \textbf{84.44} & 32.15 & \textbf{37.06} & \textbf{99.91} \\
            \bottomrule
        \end{tabular}
    }
    \vspace{-5mm}
    \label{tab:visionflan_appendix}
\end{table*}

\paragraph{Superiority across Data Scales.}
As shown in Figure~\ref{fig:teaser}, DICS consistently outperforms all methods across varying sampling ratios. In low-data ratios ($<$25\%), it establishes a distinct lead, validating that our consistency metrics effectively identify high-value samples. 
Notably, DICS achieves peak performance at the 50\% ratio (103.15\%).
While extending to high-data regimes ($>$75\%) leads to generic performance saturation or slight decline across all methods, DICS maintains significantly greater stability and resilience. 
In contrast, baselines like PRISM~\cite{bi2025prism} experience sharper degradation due to the inclusion of low quality samples. This confirms DICS's ability to maximize data efficiency while effectively filtering out inconsistent samples.
Notably, $p$ is not fixed but depends on the specific metric and dataset; other sample-level methods also exhibit such peaks (e.g., 75\% for PRISM and DataTailor in Figure~\ref{fig:teaser}).

\begin{figure}[h]
    \centering
    \includegraphics[width=1.0\linewidth]{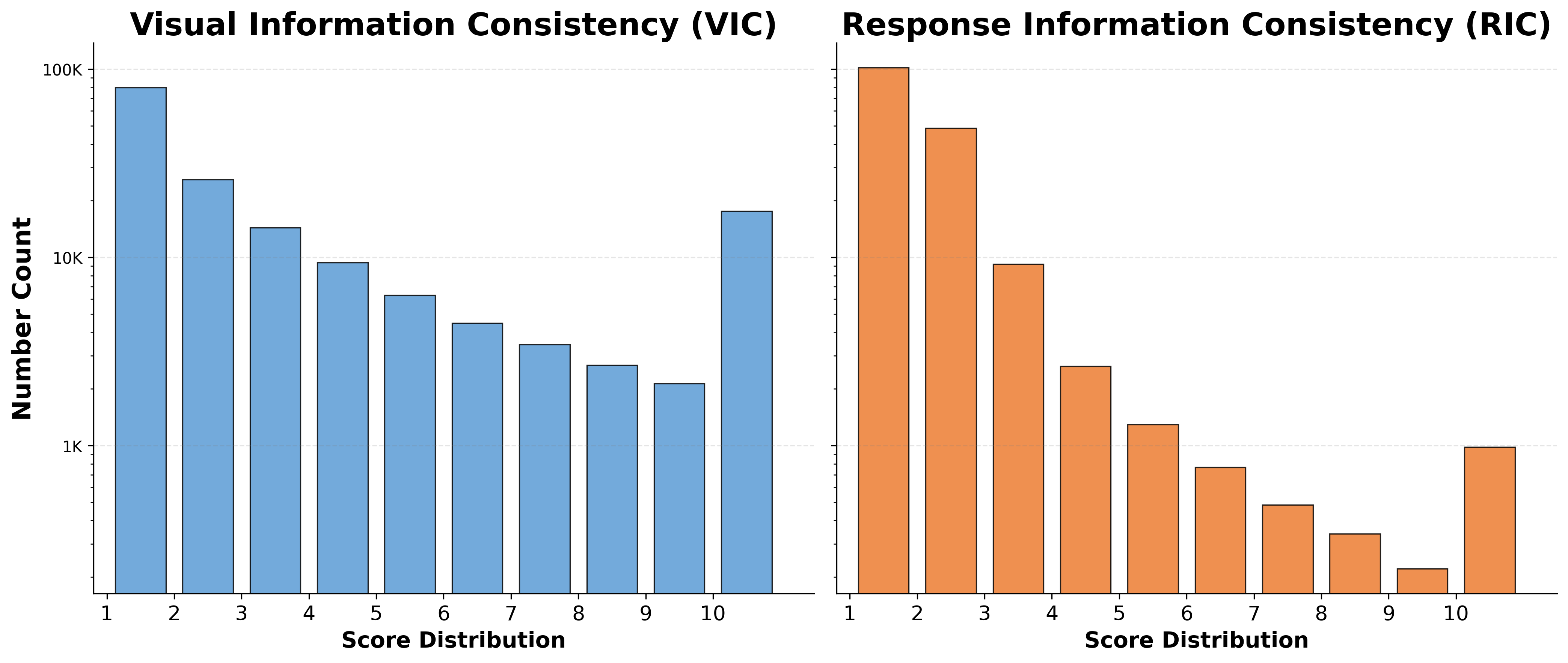}
    \caption{\textbf{The distribution of DIC scores of 25\% selected data of LLaVA-1.5-665K dataset.}}
    \label{fig:vic_distribution}
\end{figure}

\paragraph{Distributions of DIC Scores.}
\label{sec:appendix:distribution}
Figure~\ref{fig:vic_distribution} illustrates the distribution of VIC and RIC scores within the selected 25\% subset of LLaVA-1.5-665K dataset. The samples are concentrated in the higher value ranges for both metrics, confirming that our method effectively prioritizes data with strong visual dependency and logical coherence. Simultaneously, the distribution exhibits a broad coverage across these high scores rather than collapsing into a narrow peak. This healthy variance indicates that samples of varying complexity are retained, effectively preventing homogeneity. This balance ensures that the final dataset achieves high intrinsic consistency without sacrificing the distributional diversity required for robust model generalization.

\paragraph{Statistical Analysis.} 
To further validate the statistical significance and robustness of our proposed framework, we report the mean scores and standard deviations over 5 independent runs for the main experiments corresponding to Table~\ref{tab:main} and Table~\ref{tab:visionflan}. The detailed results are summarized in Table~\ref{tab:std_results}. As shown in the table, the variances across all methods are extremely minimal (all standard deviations $\le 0.13$). Specifically, the performance gaps between our DICS framework and the strongest baseline, PRISM (e.g., $+1.79$ in Table~\ref{tab:main} and $+0.63$ in Table~\ref{tab:visionflan}), are substantially larger than the standard deviations. This confirms that our improvements are statistically significant and clearly exceed random fluctuations. This overall stability thoroughly demonstrates our method's robustness and proves that the performance gains are intrinsic to our data selection strategy.

\begin{table}[t]
    \centering
    \caption{\textbf{Mean scores and standard deviations over 5 independent runs.} The results correspond to the overall performance in Table~\ref{tab:main} and Table~\ref{tab:visionflan}.}
    \small 
    \begin{tabular}{l cc}
        \toprule
        \multirow{2}{*}{\textbf{Method}} & \multicolumn{2}{c}{\textbf{Score (Mean $\pm$ Std)}} \\
        \cmidrule(lr){2-3}
        & \textbf{Table~\ref{tab:main}} & \textbf{Table~\ref{tab:visionflan}} \\
        \midrule
        Full Dataset & 46.29 $\pm$ 0.03 & 43.47 $\pm$ 0.06 \\
        \midrule
        Random Selection & 44.64 $\pm$ 0.13 & 41.85 $\pm$ 0.07 \\
        DataTailor & 44.86 $\pm$ 0.10 & 41.94 $\pm$ 0.07 \\
        PRISM & 45.18 $\pm$ 0.05 & 42.90 $\pm$ 0.03 \\
        \rowcolor{cyan!15} 
        DICS (Ours) & \textbf{46.97 $\pm$ 0.04} & \textbf{43.53 $\pm$ 0.05} \\
        \bottomrule
    \end{tabular}
    \vspace{-5pt}
    \label{tab:std_results}
\end{table}

\subsection{Robustness and Scalability} 
\begin{table*}[t]
    \centering
    \caption{\textbf{Detailed generalization analysis across architectures and scales.} This table provides the granular results for the cross-architecture experiments for Table~\ref{tab:generalization_matrix}. \textit{Model} indicates the backbone being trained, while \textit{Selector} indicates the model used to calculate DIC scores for data filtering. All subset selection methods employ a 25\% sampling ratio. The best results for each target model are highlighted in \textbf{bold}, and second-best are \underline{underlined}.}
    \resizebox{\textwidth}{!}{
        \begin{tabular}{l l ccc ccc ccc cc c c}
            \toprule
            & & \multicolumn{3}{c}{\textbf{General Capability}} 
            & \multicolumn{3}{c}{\textbf{Knowledge \& Reasoning}} 
            & \multicolumn{3}{c}{\textbf{Document Understanding}} 
            & \multicolumn{2}{c}{\textbf{Hallucination}} 
            & \multicolumn{1}{c}{\textbf{Dialogue}} 
            & \\
            \cmidrule(lr){3-5} \cmidrule(lr){6-8} \cmidrule(lr){9-11} \cmidrule(lr){12-13} \cmidrule(lr){14-14}
            
            \textbf{Model} & \textbf{Selector}
            & \textbf{MME} & \multicolumn{2}{c}{\textbf{MMBench}} 
            & \textbf{MMMU} & \textbf{SQA-I} & \textbf{VizWiz} 
            & \textbf{AI2D} & \textbf{DocVQA} & \textbf{InfoVQA} 
            & \textbf{POPE} & \textbf{Hallusion} 
            & \textbf{MMVet} & \textbf{Rel. (\%)} \\
            
            & & & \textbf{en} & \textbf{cn} & & & & & & & & \textbf{Bench} & & \\
            \midrule
            
            & Full Dataset
            & 59.92 & 60.22 & 51.93 & 35.89 & \textbf{68.32} & 31.79 & 47.91 & 23.82 & 22.30 & 84.49 & 27.80 & 41.01 & 100.00 \\
            
            & Random Selection
            & 59.28 & 56.27 & 51.93 & 35.56 & 66.98 & 29.53 & 50.91 & 21.62 & 21.63 & 81.58 & \textbf{29.46} & 38.07 & 97.74 \\
            
            \rowcolor{cyan!15} \cellcolor{white}
            & DICS (LLaVA-1.5-7B)
            & \textbf{63.39} & \textbf{59.52} & \textbf{55.03} & \textbf{36.11} & 66.14 & \textbf{31.29} & \textbf{50.52} & \textbf{24.64} & \textbf{23.11} & \textbf{84.91} & 26.79 & 41.74 & \textbf{101.40} \\
            
            \rowcolor{cyan!15} \cellcolor{white}
            \multirow{-4}{*}{\textbf{LLaVA-1.5-7B}} 
            & DICS (Qwen2-VL-7B)
            & \underline{62.27} & \underline{58.98} & \underline{54.57} & \underline{36.00} & 65.54 & \underline{31.22} & 48.87 & \underline{24.24} & \underline{23.17} & 83.78 & 25.19 & \textbf{42.48} & \underline{100.16} \\

            \midrule

            & Full Dataset
            & 78.06 & 78.72 & 78.02 & 48.11 & 81.90 & 41.88 & 80.83 & 91.92 & 73.14 & 87.97 & \textbf{47.18} & 57.84 & 100.00 \\
            
            & Random Selection
            & \textbf{79.51} & 78.72 & 76.70 & 46.67 & 81.51 & 41.87 & \underline{81.28} & \underline{92.72} & 74.27 & 86.37 & \underline{46.64} & \textbf{61.10} & 100.21 \\
            
            \rowcolor{cyan!15} \cellcolor{white}
            & DICS (LLaVA-1.5-7B)
            & \underline{78.17} & \textbf{79.10} & \textbf{79.72} & \textbf{52.00} & 82.10 & \textbf{42.28} & \textbf{81.28} & 92.32 & \textbf{74.50} & \textbf{88.51} & 45.15 & \underline{61.01} & \textbf{101.25} \\
            
            \rowcolor{cyan!15} \cellcolor{white}
            \multirow{-4}{*}{\textbf{Qwen2-VL-7B}}
            & DICS (Qwen2-VL-7B)
            & 77.39 & \underline{78.64} & \underline{78.25} & \textbf{52.00} & \textbf{83.09} & 41.73 & 81.02 & \textbf{92.37} & 73.79 & \underline{87.68} & \underline{46.64} & 60.64 & \underline{100.91} \\

            \cmidrule{2-15}
            \textit{Ref: Qwen2-VL-7B-Base}
            & -
            & 67.94 & 75.77 & 73.99 & 50.33 & 81.71 & 1.94 & 79.34 & 4.75 & 0.78 & 78.98 & 42.42 & 54.68 & 72.45 \\

            \textit{Ref: Qwen2-VL-7B-Instruct}
            & - 
            & 81.80 & 79.18 & 79.49 & 47.00 & 84.23 & 42.46 & 82.51 & 93.83 & 75.51 & 85.82 & 50.58 & 63.94 & 102.46 \\

            \midrule
            
            & Full Dataset
            & 54.06 & \textbf{62.93} & 56.81 & \textbf{37.33} & 69.16 & \textbf{32.83} & \textbf{58.13} & \underline{26.39} & \textbf{26.16} & 81.96 & 24.80 & 44.17 & 100.00 \\
            
            & Random Selection
            & \underline{62.44} & 60.68 & 53.25 & 36.56 & \textbf{71.84} & 30.97 & 55.99 & 24.21 & 25.19 & \underline{85.16} & 23.44 & \textbf{44.59} & 99.93 \\
            
            \rowcolor{cyan!15} \cellcolor{white}
            \multirow{-3}{*}{\textbf{LLaVA-1.5-13B}}
            & DICS (LLaVA-1.5-7B)
            & \textbf{63.62} & \underline{61.76} & \textbf{58.05} & \underline{37.11} & \underline{70.10} & \underline{32.06} & \underline{56.02} & 26.04 & \underline{25.90} & \textbf{85.93} & \textbf{24.92} & 43.49 & \textbf{101.79} \\
            
            \bottomrule
        \end{tabular}
    }
    \vspace{-5pt}
    \label{tab:detailed_arch_scale}
\end{table*}
Based on the granular results presented in Table~\ref{tab:detailed_arch_scale}, we provide a deeper analysis of the effectiveness of DICS across different model architectures and scales.

\paragraph{Architecture-Agnostic DIC Metrics.}
A key question in data selection is whether the filtered subset is biased towards the selector model's specific preferences or if it captures intrinsic data quality. 
Observing the \textbf{Qwen2-VL-7B} target experiments, we find that data selected by the \textbf{LLaVA-1.5-7B} model achieves a relative performance of \textbf{101.25\%}, slightly outperforming the subset selected by Qwen2-VL-7B itself (100.91\%) and significantly surpassing the full dataset baseline. 
This indicates that DICS captures \textit{universal} consistency  between visual and textual modalities rather than model-specific patterns. The high transferability implies that we can utilize a single, well-established model (like LLaVA) to curate data for different architectures without re-computing scores for each new target model.

\paragraph{Scalability via Weak-to-Strong Generalization.}
The results on the \textbf{LLaVA-1.5-13B} target model offer compelling evidence for the scalability of our approach. Training the 13B model using data selected by the much smaller \textbf{7B} selector yields a relative performance of \textbf{101.79\%}, the highest improvement margin among all settings. 
This "Weak-to-Strong" generalization capability is practically significant. It suggests that DICS enables a computationally efficient pipeline where lightweight models act as "data filters" for training large-scale models. This significantly reduces the computational overhead of the data selection phase, which is traditionally a bottleneck when processing massive datasets. Consequently, this transferability suggests that well-established models (like LLaVA) have the potential to serve as efficient data curators for diverse architectures, reducing the computational redundancy of re-calculating scores for every new target model.

\paragraph{Architectural Generalizability of DICS.}
To strictly validate the quality of our selected subsets, we compare our models (fine-tuned from \textbf{Qwen2-VL-7B-Base}) against the official checkpoints. 
As shown in the table~\ref{tab:detailed_arch_scale}, the Qwen2-VL-7B-Base model exhibits limited instruction-following capabilities (Score: 51.05), serving as a lower-bound baseline. 
However, by fine-tuning on just 25\% of the LLaVA-665K dataset selected by DICS, our model achieves a substantial performance leap to \textbf{71.35}. 
Remarkably, this result bridges the gap towards the official \textbf{Qwen2-VL-7B-Instruct} model (Score: 72.20) to within a 1\% margin. This demonstrates that prioritizing data intrinsic consistency allows for highly efficient model training with minimal resource consumption.

\subsection{Ablation Studies and Analysis} 
\begin{table*}[t]
    \centering
    \caption{\textbf{Ablation study on Data Intrinsic Consistency (DIC).} We analyze the impact of individual consistency metrics, different fusion strategies, and token weighting. All methods use the LLaVA-1.5-665K dataset with a 25\% sampling ratio on the LLaVA-1.5-7B model. The best results are highlighted in \textbf{bold}.}
    \resizebox{\textwidth}{!}{
        \begin{tabular}{l ccc ccc ccc cc c c}
            \toprule
            & \multicolumn{3}{c}{\textbf{General Capability}} 
            & \multicolumn{3}{c}{\textbf{Knowledge \& Reasoning}} 
            & \multicolumn{3}{c}{\textbf{Document Understanding}} 
            & \multicolumn{2}{c}{\textbf{Hallucination}} 
            & \multicolumn{1}{c}{\textbf{Dialogue}} 
            & \\
            \cmidrule(lr){2-4} \cmidrule(lr){5-7} \cmidrule(lr){8-10} \cmidrule(lr){11-12} \cmidrule(lr){13-13}
            
            \textbf{Metric Variant} 
            & \textbf{MME} & \multicolumn{2}{c}{\textbf{MMBench}} 
            & \textbf{MMMU} & \textbf{SQA-I} & \textbf{VizWiz} 
            & \textbf{AI2D} & \textbf{DocVQA} & \textbf{InfoVQA} 
            & \textbf{POPE} & \textbf{Hallusion} 
            & \textbf{MMVet} & \textbf{Rel. (\%)} \\
            
            & & \textbf{en} & \textbf{cn} & \textbf{(val)} & & & & & & & \textbf{Bench} & & \\
            \midrule

            \multicolumn{13}{l}{\textit{\textbf{Baselines}}} \\ 
            Full Dataset
            & 59.92 & \textbf{60.22} & 51.93 & 35.89 & \textbf{68.32} & \textbf{31.79} & 47.83 & 23.82 & 22.30 & 84.49 & 27.80 & 41.01 & 100.00\% \\
            FPS (Diversity Only)
            & 58.51 & 54.72 & 49.23 & 33.67 & 66.58 & 29.94 & 50.45 & 22.78 & 22.35 & 84.08 & 27.44 & 39.08 & 97.02\% \\
            \midrule
            
            \multicolumn{13}{l}{\textit{\textbf{Single Consistency Metric}}} \\ 
            w/ VIC only
            & \textbf{63.87} & 59.06 & 54.18 & \textbf{36.44} & 66.63 & 31.50 & \textbf{50.97} & 24.24 & 22.96 & 83.92 & 27.72 & 39.54 & 101.02\% \\
            
            w/ RIC only
            & 59.63 & 53.17 & 48.92 & 35.44 & 66.09 & 30.22 & 49.68 & 22.38 & 22.29 & \textbf{85.83} & \textbf{28.13} & 40.28 & 97.60\% \\
            \midrule
            
            \multicolumn{13}{l}{\textit{\textbf{Metric Fusion Strategies}}} \\ 
            Sum (VIC + RIC)
            & 59.01 & 57.12 & 53.17 & 33.33 & 66.78 & 31.31 & 48.64 & 24.10 & 21.75 & 84.67 & 26.22 & 41.56 & 98.61\% \\
            
            Product (VIC $\times$ RIC)
            & 58.12 & 56.66 & 53.95 & 35.89 & 66.68 & 31.02 & 48.06 & 24.25 & 22.85 & 84.70 & 27.63 & 39.17 & 98.84\% \\
            \midrule
            
            \multicolumn{13}{l}{\textit{\textbf{Weighted Token Strategy}}} \\ 
            DIC w/o weighted token.
            & 61.02 & 58.59 & 54.95 & 34.33 & 66.29 & 31.17 & 50.58 & 24.60 & \textbf{23.38} & 84.35 & 27.49 & 40.05 & 100.25\% \\
            \midrule
            
            \rowcolor{cyan!15} 
            \textbf{DIC (ours)}
            & 63.39 & 59.52 & \textbf{55.03} & 36.11 & 66.14 & 31.29 & 50.52 & \textbf{24.64} & 23.11 & 84.91 & 26.79 & \textbf{41.74} & \textbf{101.40\%} \\
            
            \bottomrule
        \end{tabular}
    }
    \vspace{-5pt}
    \label{tab:ablation_metrics}
\end{table*}
\begin{table*}[t]
    \centering
    \caption{\textbf{Ablation study on selection strategies.} We compare different diversity-aware selection methods applied to the high-quality subset filtered by DIC. All methods use the LLaVA-1.5-665K dataset with a 25\% sampling ratio on the LLaVA-1.5-7B model.  The best results are highlighted in \textbf{bold}.}
    \resizebox{\textwidth}{!}{
        \begin{tabular}{l ccc ccc ccc cc c c}
            \toprule
            & \multicolumn{3}{c}{\textbf{General Capability}} 
            & \multicolumn{3}{c}{\textbf{Knowledge \& Reasoning}} 
            & \multicolumn{3}{c}{\textbf{Document Understanding}} 
            & \multicolumn{2}{c}{\textbf{Hallucination}} 
            & \multicolumn{1}{c}{\textbf{Dialogue}} 
            & \\
            \cmidrule(lr){2-4} \cmidrule(lr){5-7} \cmidrule(lr){8-10} \cmidrule(lr){11-12} \cmidrule(lr){13-13}
            
            \textbf{Sampling Strategy} 
            & \textbf{MME} & \multicolumn{2}{c}{\textbf{MMBench}} 
            & \textbf{MMMU} & \textbf{SQA-I} & \textbf{VizWiz} 
            & \textbf{AI2D} & \textbf{DocVQA} & \textbf{InfoVQA} 
            & \textbf{POPE} & \textbf{Hallusion} 
            & \textbf{MMVet} & \textbf{Rel. (\%)} \\
            
            & & \textbf{en} & \textbf{cn} & & & & & & & & \textbf{Bench} & & \\
            \midrule
            
            Full Dataset
            & 59.92 & \textbf{60.22} & 51.93 & 35.89 & \textbf{68.32} & 31.79 & 47.83 & 23.82 & 22.30 & 84.49 & 27.80 & 41.01 & 100.00\% \\
            \midrule
            
            K-Means ($K$=5,000)
            & 61.58 & 58.44 & 54.72 & 35.89 & 65.94 & \textbf{32.02} & \textbf{50.87} & 24.44 & \textbf{23.54} & 84.59 & 25.62 & 40.37 & 100.47\% \\
            
            Hierarchical
            & \textbf{64.09} & 57.20 & 53.79 & \textbf{36.44} & 67.48 & 31.91 & 49.90 & \textbf{24.77} & 23.15 & 83.97 & \textbf{29.22} & 41.19 & 101.39\% \\
            \midrule
            
            \rowcolor{cyan!15} 
            \textbf{FPS (Ours)}
            & 63.39 & 59.52 & \textbf{55.03} & 36.11 & 66.14 & 31.29 & 50.52 & 24.64 & 23.11 & \textbf{84.91} & 26.79 & \textbf{41.74} & \textbf{101.40\%} \\
            
            \bottomrule
        \end{tabular}
    }
    \vspace{-5pt}
    \label{tab:ablation_sampling}
\end{table*}

\paragraph{Effectiveness of DIC.}
We dissect the contribution of each DIC component to understand their roles in data selection in Table~\ref{tab:ablation_metrics}.
\begin{itemize}
    \item \textbf{Necessity of Quality Filtering:} The "Diversity Only" baseline (selecting data solely based on feature embedding distance) yields the lowest performance (97.02\%). This confirms that maximizing diversity without ensuring intrinsic quality introduces low-quality samples that degrade model capabilities.
    \item \textbf{Complementarity of VIC and RIC:} VIC serves as the primary driver for general capabilities, achieving a high relative score of 101.02\%. In contrast, while RIC alone underperforms in general tasks (97.60\%), it excels in hallucination benchmarks like \textbf{POPE} and \textbf{HallusionBench}. Consequently, VIC ensures the fidelity of visual and textual content, whereas RIC promotes structural and logical consistency, effectively reducing the incidence of hallucinations in complex multimodal tasks.
    \item \textbf{Fusion Strategies:} Simple fusion strategies like Sum or Product fail to outperform the single VIC metric. However, our proposed \textbf{min(VIC, RIC)} operator achieves the best overall performance (101.40\%). The "min" operator acts as a strict gate, requiring samples to be high quality in \textit{both} visual alignment and response logic, effectively filtering out partial outlier samples.
    \item \textbf{Impact of Token Weighting:} Removing the POS-based weighting leads to a performance drop (101.40\% $\rightarrow$ 100.25\%), validating that focusing on content-rich tokens (nouns, verbs, etc) provides a more accurate assessment of semantic consistency.
\end{itemize}

\paragraph{Impact of Selection Strategies.}
Once the high-quality candidate set is identified, the choice of selecting strategy for the second stage is critical for maintaining distribution coverage. We detail the results in Table~\ref{tab:ablation_sampling}. As shown in Table \ref{tab:ablation_sampling}, all diversity-aware strategies (Hierarchical, and FPS) consistently outperform the full dataset baseline (all $>$100\%). This robustness confirms that the core contribution comes from the DIC-based quality filtering stage, which successfully identifies a high-value data distribution regardless of the subsequent selection method.

To further validate the effectiveness of DIC, we introduce two direct validation experiments.

\paragraph{Controlled Corruption Analysis.} 
We randomly sample 1k examples from LLaVA-1.5-665K and recompute the VIC/RIC scores after applying three types of controlled corruption to each sample. To quantitatively assess their sensitivity to data-quality degradation, we examine the proportion of samples whose scores decrease after corruption, the median relative change in scores, and the Area Under the ROC Curve (AUC). The AUC measures the separability between the score distributions of the original and corrupted samples; an AUC of 0.5 indicates that the two distributions are completely indistinguishable, while values closer to 1 indicate stronger separability. The three types of corruption are defined as follows:
\begin{itemize}
    \item \textbf{Image-text mismatch:} We randomly replace the original image with an unrelated one.
    \item \textbf{Injected factual errors / hallucinations:} We use a strong VLM, Qwen3.5-397B-A17B, to modify the answer by deliberately injecting hallucinations or incorrect facts.
    \item \textbf{Generic / irrelevant responses:} We randomly replace the answer with a templated generic response or an irrelevant response.
\end{itemize}

\begin{table}[htbp]
    \centering
    \small 
    \caption{Evaluation metrics under controlled corruption analysis.}
    \label{tab:corruption_analysis}
    \resizebox{\columnwidth}{!}{
    \begin{tabular}{l c c c c c c c}
        \toprule
        \textbf{Corruption Type} & \textbf{Metric} & \textbf{Orig.} & \textbf{Corr.} & \textbf{Diff.} & \textbf{Rel. $\Delta$} & \textbf{Decr. \%} & \textbf{AUC} \\
        \midrule
        Image-text mismatch & VIC & 1.827 & 0.985 & $-$0.811 & $-$43.0\% & 89.0\% & 0.816 \\
        Injected errors / halluc. & VIC & 1.827 & 1.135 & $-$0.607 & $-$32.8\% & 77.7\% & 0.689 \\
        Generic / irrelevant & RIC & 1.778 & 1.424 & $-$0.225 & $-$13.1\% & 75.6\% & 0.701 \\
        \bottomrule
    \end{tabular}
    }
    \vspace{-5pt} 
\end{table}

As shown in Table~\ref{tab:corruption_analysis}, all three corruption types lead to a significant drop in the corresponding scores, demonstrating the high sensitivity of DIC to data-quality degradation.

\paragraph{Human and VLM-as-Judge Quality Assessment.} 
We sample 750 examples from LLaVA-1.5-665K, with 250 examples drawn from each of the low-, medium-, and high-DIC quantile groups. A strong VLM, Qwen3.5-397B-A17B, serving as the judge model, and human annotators independently rate the overall quality of each sample on a 1--5 scale, where 1 indicates the lowest quality and 5 indicates the highest quality. The ratings jointly consider visual dependency, instruction relevance, and the degree of hallucination.

As shown in Table~\ref{tab:human_vlm_judge}, the quality scores from both evaluation protocols increase with DIC. The Spearman correlations between DIC and the quality scores are $\rho=0.20$ for the VLM judge and $\rho=0.27$ for human evaluation, with $p<0.001$. These results indicate that DIC is broadly aligned with both human and strong-model quality judgments, and can serve as an effective and scalable signal for data quality assessment.

\begin{table}[htbp]
    \centering
    \caption{Quality scores across different DIC quantiles evaluated by VLM-as-Judge and human annotators.}
    \label{tab:human_vlm_judge}
    \resizebox{\columnwidth}{!}{
    \begin{tabular}{l c c c c}
        \toprule
        \textbf{DIC Quantile} & \textbf{Samples} & \textbf{Median DIC} & \textbf{Mean VLM} & \textbf{Mean Human} \\
        \midrule
        Low & 250 & 0.93 & 2.59 & 2.58 \\
        Medium & 250 & 1.28 & 2.72 & 2.92 \\
        High & 250 & 2.55 & 2.86 & 3.17 \\
        \bottomrule
    \end{tabular}
    }
\end{table}

\begin{figure*}[t!]
    \centering
    \includegraphics[width=1\linewidth]{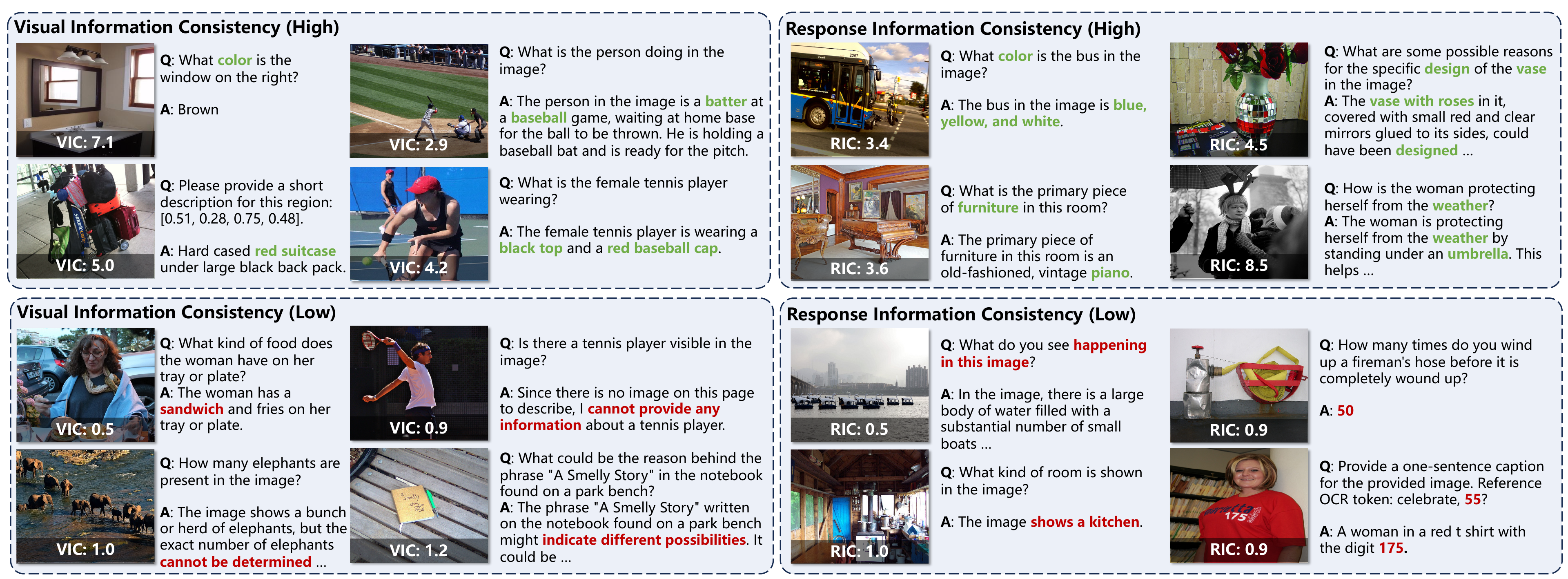}
    \caption{\textbf{More visualization of DIC scores.}}
    \label{fig:dic_vis_appendix}
\end{figure*}

\paragraph{Visualization of VIC and RIC.}
To provide a more intuitive understanding of our proposed metric, we present representative samples with high and low DIC scores from the LLaVA-1.5-665K dataset in Figure~\ref{fig:dic_vis_appendix}. 
Samples with high Visual Information Consistency (VIC) scores are often characterized by distinct visual elements, well-defined questions and informative answers, demonstrating a strong cross-modal alignment between vision and language. Conversely, instances with low VIC scores often contain ambiguous or incorrect answers, where the visual input provides minimal information gain for the given question. 
Regarding Response Information Consistency (RIC), high-scoring samples generally exhibit rich descriptive content, clear reasoning logic, and a strong semantic correlation between the question and the answer. In contrast, samples with low RIC scores tend to produce vague, generic responses that lack informative content.

\paragraph{Qualitative Comparison via Attention Map.}
\label{sec:appendix:attention_analysis}
To generate the text-to-image attention maps, we perform a complete forward pass of the LLaVA-1.5 model (which consists of 32 layers in total) and extract the attention weights from a specific layer of the language model (specifically, layer 21). We empirically select this intermediate-deep layer because it offers an optimal balance between spatial precision and semantic relevance. In previous studies and our observations, deeper layers tend to capture more abstract semantic information but lack fine-grained spatial localization, whereas shallower layers produce more diffuse and noisy attention patterns. We aggregate the attention weights from the generated response tokens to the visual patch tokens, yielding a 576-dimensional text-to-image attention distribution. This distribution is then reshaped into a $24 \times 24$ spatial grid, smoothed, and upsampled to the original image resolution using bilinear interpolation. Finally, we apply contrast enhancement and the JET colormap to generate the heatmap, which is overlaid onto the original image. The baseline for comparison is the model trained on a 15\% randomly selected subset.

\begin{figure*}[h!]
    \centering
    \includegraphics[width=1\linewidth]{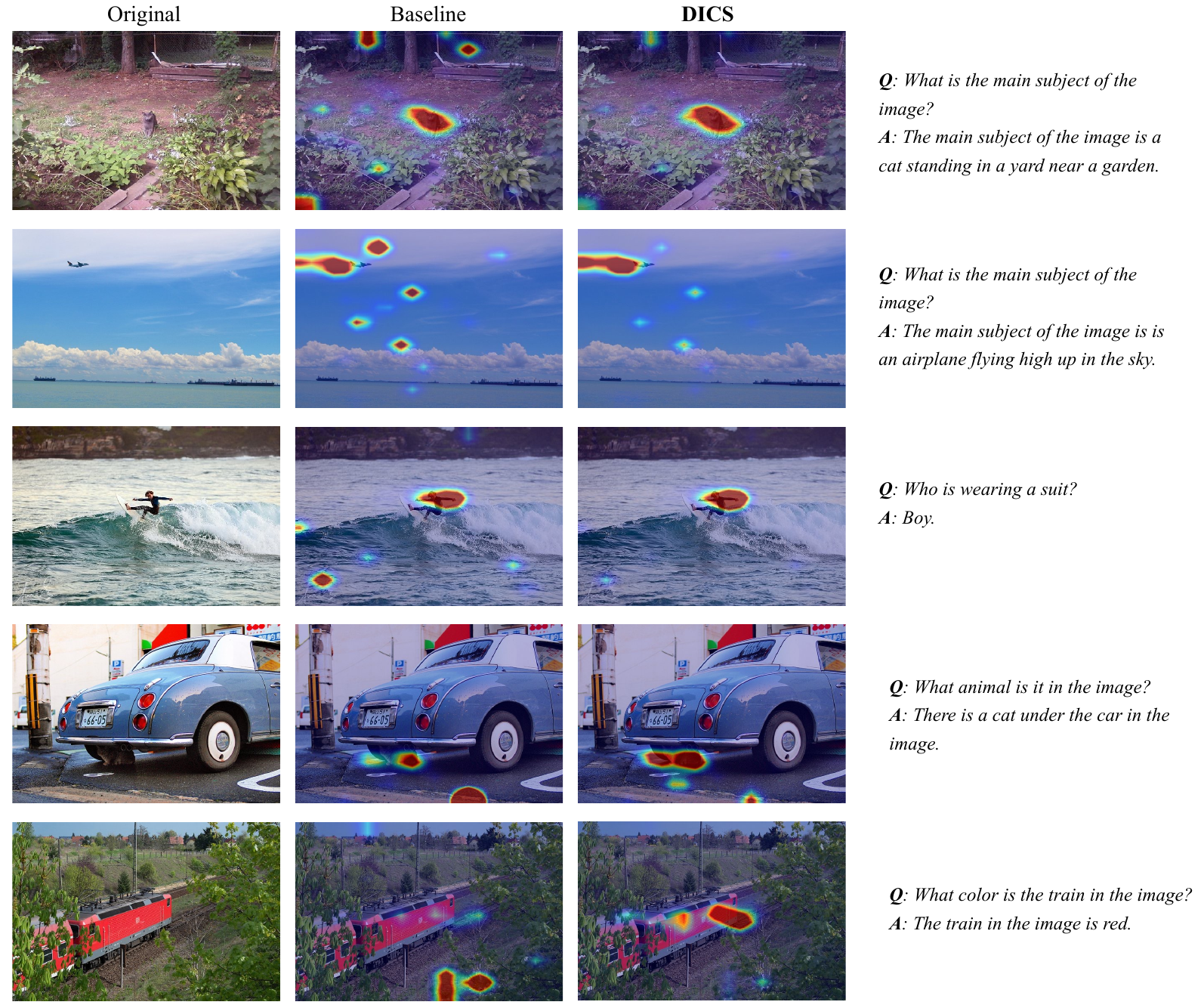}
    \caption{\textbf{Analysis of attention.} Compared to the random selection baseline, DICS encourages the model to pay more attention to the core visual details, rather than irrelevant background information.}
    \label{fig:attn_map_appendix}
\end{figure*}

As illustrated in Figure~\ref{fig:attn_map_appendix}, the model trained with our DICS framework demonstrates highly concentrated and accurate attention on the core target objects. For instance, when answering questions about specific visual details, the DICS model precisely localizes the relevant entities. In contrast, the random selection baseline often distributes its attention diffusely across the image or focuses on irrelevant background regions. This qualitative evidence further substantiates that DICS effectively encourages the model to rely on critical visual evidence rather than language priors.

\subsection{Computational Efficiency}
\label{sec:appendix:compute_cost}
\begin{table}[t!]
\centering
\caption{\textbf{Computational cost analysis on LLaVA-665K ($N$=665K) using 8$\times$A100 GPUs.} The pipeline is dominated by inference-time costs ($O(N\mathcal{F})$), while the selection overhead is minimal.}
\label{tab:computation_cost}
\resizebox{1.0\linewidth}{!}{
\begin{tabular}{lccc}
\toprule
\textbf{Stage} & \textbf{Complexity} & \textbf{Time} & \textbf{Type} \\
\midrule
Warmup Training & $O(EN\mathcal{F})$ & $<$1h & Gradient/Fwd \\
DIC \& Embedding & $O(N\mathcal{F})$ & $\sim$10h & Inference Only \\
Data Selection & $O(N\log N + MND)$ & $<$1h & Vector Ops \\
\bottomrule
\end{tabular}
}
\vspace{-3mm}
\end{table}

\begin{table}[htbp]
    \centering
    \caption{Computational cost comparison (in hours) among different methods.}
    \label{tab:efficiency}
    
    \resizebox{\columnwidth}{!}{
    \begin{tabular}{l c c c}
        \toprule
        \textbf{Method} & \textbf{Data Selection} & \textbf{Visual Instruction Tuning} & \textbf{Overall} \\
        \midrule
        Full-Finetuning & - & 15.0 & 15.0 \\
        COINCIDE (EMNLP 2024) & 16.3 & 2.0 & 18.3 \\
        DataTailor (ICCV 2025) & 14.5 & 2.0 & 16.5 \\
        \textbf{DICS (Ours)} & \textbf{12.0} & \textbf{2.0} & \textbf{14.0} \\
        \bottomrule
    \end{tabular}
    }
\end{table}
We analyze the computational efficiency of our data selection pipeline. Let $\mathcal{F}$ denote the complexity of a single forward pass. Given a dataset of $N$ samples, a target subset size $M$, and embedding dimension $D$, the pipeline's computational cost scales linearly with $N$. As summarized in Table~\ref{tab:computation_cost}, this linear scaling ensures high scalability. The detailed breakdown is as follows:

\begin{itemize}
    \item \textit{Warmup Training:} We perform parameter-efficient LoRA fine-tuning for $E$ epochs. Since LoRA updates only a fraction of parameters and the number of epochs is small, the complexity $O(EN\mathcal{F})$ remains computationally lightweight.

    \item \textit{Metric \& Embedding Computation:} To maximize efficiency, we integrate embedding extraction with metric computation. The visual and text embeddings are retrieved from the hidden states during the same forward passes used for VIC/RIC scores. This concurrent execution eliminates redundant model inference, maintaining a linear complexity of $O(N\mathcal{F})$ with negligible marginal cost for embedding storage. 
    
    \item \textit{Data Selection:} This stage involves score-based filtering ($O(N\log N)$) and Farthest Point Sampling ($O(MND)$). As $M \ll N$ and these operations involve only lightweight vector operations (e.g., matrix multiplication) without model invocation, the overhead here is negligible, allowing for rapid execution even on standard CPUs.
\end{itemize}

\end{document}